\PassOptionsToPackage{table}{xcolor}
\documentclass[letterpaper]{article}
\usepackage[preprint]{aaai2027}
\usepackage[hyphens]{url}
\usepackage{graphicx}
\usepackage{natbib}
\usepackage{caption}
\usepackage{algorithm}
\usepackage{algorithmic}
\usepackage{booktabs}
\usepackage{multirow}
\usepackage{placeins}
\usepackage{amsmath}
\usepackage{amssymb}
\newif\ifvbenchresults
\newif\ifdensitybthreeablations
\newif\ifdensitybthreemechanisms
\newif\ifdensitybthreeheadwise
\providecommand{\includesupplementarypreview}{1}
\IfFileExists{generated/full_vbench_30_60_120_rows.tex}{\vbenchresultstrue}{\vbenchresultsfalse}
\IfFileExists{generated/densitykv_b3_parameter60_rows.tex}{\densitybthreeablationstrue}{\densitybthreeablationsfalse}
\IfFileExists{generated/densitykv_b3_mechanism60_rows.tex}{\densitybthreemechanismstrue}{\densitybthreemechanismsfalse}
\IfFileExists{generated/densitykv_b3_headwise60_rows.tex}{\densitybthreeheadwisetrue}{\densitybthreeheadwisefalse}
\newcommand{\qualitativefigure}[2][]{\includegraphics[#1]{#2}}
\IfFileExists{figures/causal_forcing_densitykv_panda_panel.pdf}{%
    \newcommand{\teasercaption}{\textbf{DensityKV} maintains generated history in per-head KV banks whose local density growth is controlled after admission. Left: historical-attention maps show which past states each new segment uses, while admission profiles show when retained KV tokens entered the bank. Right: access to this retained history helps long videos maintain subjects, objects, and scenes more consistently over time.}%
}{%
    \IfFileExists{figures/causal_forcing_densitykv_panel.pdf}{%
        \newcommand{\teasercaption}{\textbf{DensityKV} maintains generated history in per-head KV banks whose local density growth is controlled after admission. Left: historical-attention maps show which past states each new segment uses, while admission profiles show when retained KV tokens entered the bank. Right: access to this retained history helps long videos maintain subjects, objects, and scenes more consistently over time.}%
    }{%
        \IfFileExists{figures/longlive_densitykv_panda_panel.pdf}{%
            \newcommand{\teasercaption}{\textbf{DensityKV} maintains generated history in per-head KV banks whose local density growth is controlled after admission. Left: historical-attention maps show which past states each new segment uses, while admission profiles show when retained KV tokens entered the bank. Right: access to this retained history helps long videos maintain subjects, objects, and scenes more consistently over time.}%
        }{%
            \IfFileExists{figures/self_forcing_densitykv_panel.pdf}{%
                \newcommand{\teasercaption}{\textbf{DensityKV} maintains generated history in per-head KV banks whose local density growth is controlled after admission. Left: historical-attention maps show which past states each new segment uses, while admission profiles show when retained KV tokens entered the bank. Right: access to this retained history helps long videos maintain subjects, objects, and scenes more consistently over time.}%
            }{%
                \newcommand{\teasercaption}{\textbf{DensityKV} maintains generated history in per-head KV banks whose local density growth is controlled after admission. Left: historical-key selection maps show which past states each new segment attends to; DensityKV preserves broader non-local context under the same historical KV token count. Right: access to this retained history helps long videos maintain subjects, objects, and scenes more consistently over time.}%
            }%
        }%
    }%
}
\makeatletter
\g@addto@macro\@maketitle{%
    \par%
    \begin{minipage}{\textwidth}
        \centering
        \IfFileExists{figures/causal_forcing_densitykv_panda_panel.pdf}{%
            \qualitativefigure[width=0.99\textwidth]{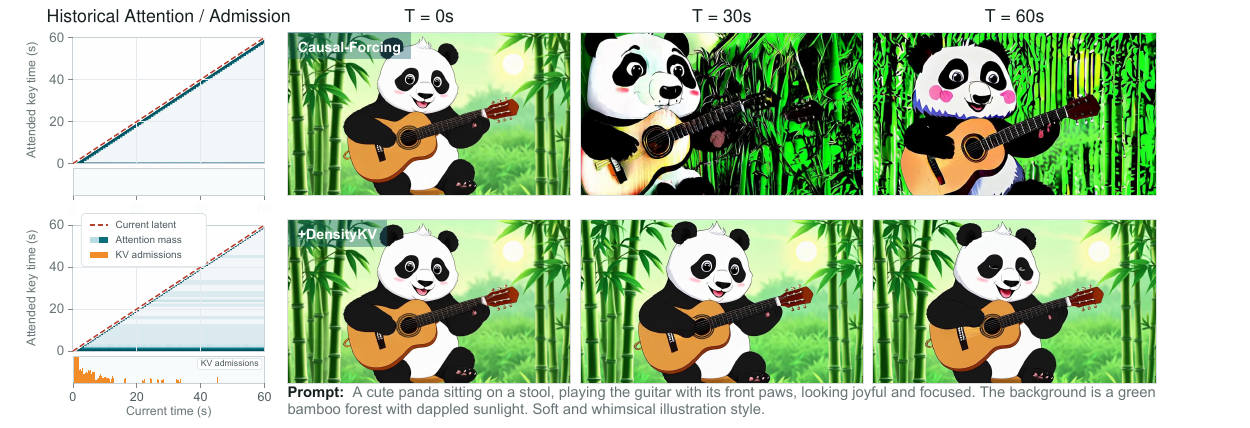}%
        }{%
            \IfFileExists{figures/causal_forcing_densitykv_panel.pdf}{%
                \qualitativefigure[width=0.99\textwidth]{figures/causal_forcing_densitykv_panel.pdf}%
            }{%
                \IfFileExists{figures/longlive_densitykv_panda_panel.pdf}{%
                    \qualitativefigure[width=0.99\textwidth]{figures/longlive_densitykv_panda_panel.pdf}%
                }{%
                    \IfFileExists{figures/self_forcing_densitykv_panel.pdf}{%
                        \qualitativefigure[width=0.99\textwidth]{figures/self_forcing_densitykv_panel.pdf}%
                    }{%
                        \qualitativefigure[width=0.99\textwidth]{figures/longlive_densitykv_panel.pdf}%
                    }%
                }%
            }%
        }
        \captionof{figure}{\teasercaption}
        \label{fig:teaser}
    \end{minipage}%
}
\makeatother

\providecommand{\pdfinfo}[1]{}
\title{DensityKV: Density-Guided KV Cache Compression for Long Video Generation}
\author{
Wenqu Zhao\textsuperscript{1}\corresponding,
Xuemin Chi\textsuperscript{1},
Xin Zhang\textsuperscript{1},
Guoqing Ma\textsuperscript{1},
Baorun Li\textsuperscript{1},\\
Jianjie Fang\textsuperscript{1,2},
Peizhi Tang\textsuperscript{1,2},
Chen Gao\textsuperscript{1,2},
Wei Wu\textsuperscript{1}
}
\affiliations{
\textsuperscript{1} Manifold AI \qquad
\textsuperscript{2} Tsinghua University\\
Code: \url{https://github.com/ZhaoWQQ/DensityKV}
}

\begin{document}
\maketitle

\begin{abstract}
Autoregressive video diffusion models enable streaming generation through sliding-window attention, but each generated block is conditioned on previously generated content, causing appearance and motion errors to propagate recursively over time. 
Historical key--value (KV) memory preserves earlier subject and scene states
and helps maintain long-horizon consistency. However, retaining every generated
state creates a historical archive that grows continuously with the rollout,
while recurrent states repeatedly add redundant coverage.
To address this problem, we propose \emph{DensityKV}, a training-free historical
KV bank management strategy.
DensityKV maintains a separate token-level KV bank for each attention head and
measures local redundancy among the post-RoPE keys that directly parameterize
attention routing using \emph{Soft-Riesz density}. By constraining neighborhood-density growth
after states enter the bank, DensityKV limits repeated historical accumulation
while preserving coherent states from each completed generation block.
Experiments across three autoregressive video generation backbones and multiple generation lengths show that, at
the same upper bound on historical KV capacity, DensityKV improves long-horizon
consistency and generation stability while keeping persistent historical
storage bounded independently of rollout length.
\end{abstract}

\section{Introduction}
Diffusion Transformers have become a strong foundation for video synthesis
\citep{peebles2023dit,wan2025}. Recent autoregressive (AR) variants turn
short-clip diffusion models into causal generators that reuse key--value (KV)
states and stream frames beyond the training horizon
\citep{yin2024causvid,huang2025selfforcing,zhu2026causalforcing,yang2025longlive}.
During long rollouts, recursive conditioning propagates appearance and motion
errors, while the sliding window that bounds attention discards earlier subject
and scene states that could stabilize later blocks.

Long-range memory can restore access to this discarded history. Existing methods
preserve early anchors, retain selected historical states, or retrieve
non-local information from previously generated content
\citep{yang2025longlive,yi2025deepforcing,zhao2026relaxforcing,hu2026longliverag}.
However, storing all historical states causes the cache to grow with generation
length, and many stored states are redundant. We study how to retain useful
native historical K/V states in a bounded bank before future queries are known.

Frame-level retention is too coarse because a frame combines repetitive
background tokens with changing appearance, pose, and motion states. A
single decision for the whole frame can either waste memory on repeated regions
or discard useful states. Token-level selection instead allows different
regions to compete for capacity according to their local coverage in key space.
Because future queries are unavailable, the selection criterion should be
query-independent and defined in the model's native attention space.

Euclidean distance between post-RoPE keys bounds their normalized logit
discrepancy for any query. We therefore use local key-space density to identify
key neighborhoods that are already well represented without observing future
queries; values are always retained or removed together with their original
keys.

\begin{figure*}[t]
    \centering
    \includegraphics[width=0.96\textwidth]{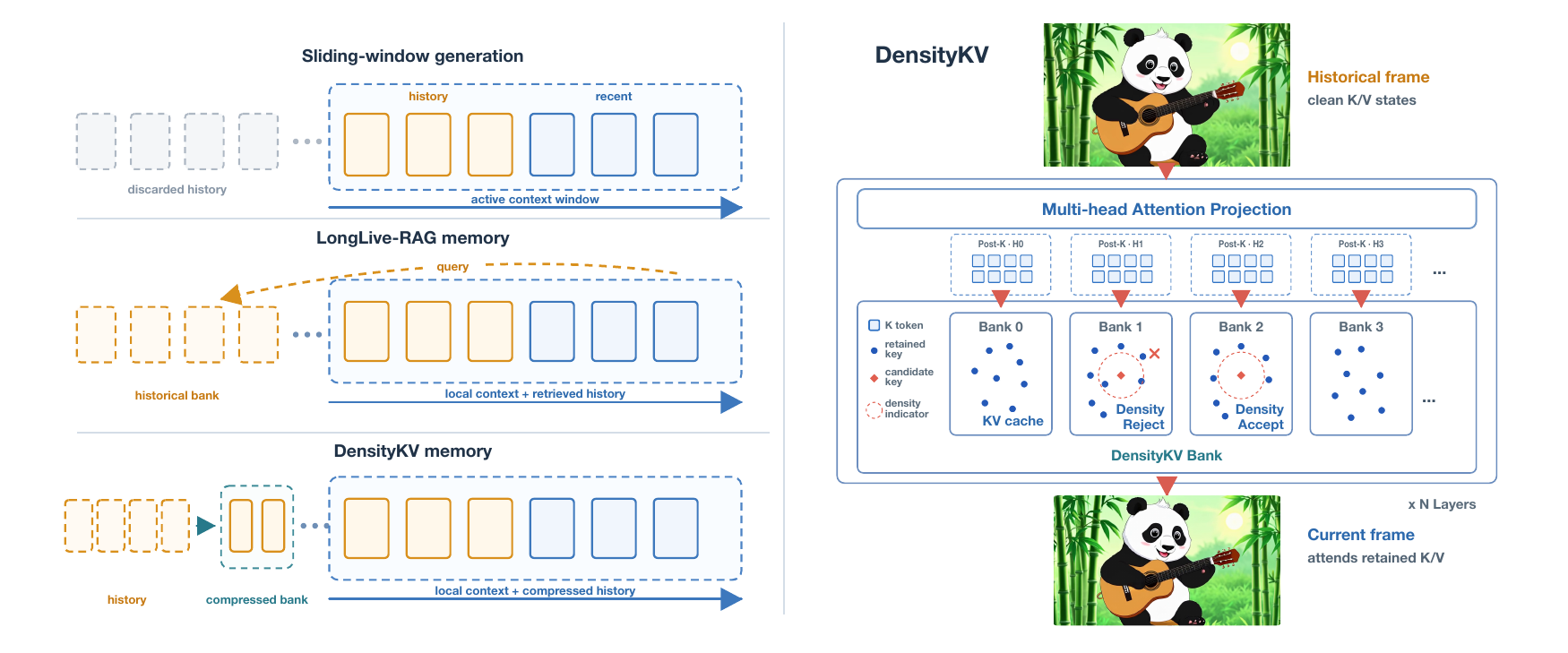}
    \caption{Overview of historical-memory construction and use. Left:
    sliding-window generation discards states outside the active context;
    LongLive-RAG stores discarded history and retrieves selected blocks;
    DensityKV instead maintains a bounded token-level history that is attended
    together with the local window. Right: at every Transformer layer, clean
    historical K/V states update independent per-head banks. Admission and
    rejection are determined by Soft-Riesz density over post-RoPE keys, while
    each value remains exactly paired with its original key. Current queries
    access the retained history through the backbone's native attention.}
    \label{fig:method-overview}
    \vspace{-5pt} 
\end{figure*}

DensityKV is a training-free historical KV bank management strategy for frozen
AR video generators. Each layer and head maintains a token-level bank of fully
denoised states. A Soft-Riesz potential measures post-RoPE key crowding, while
each state records its density at admission as a local baseline. DensityKV
admits the largest candidate prefix satisfying the relative growth constraint
and evicts states whose neighborhoods later become overrepresented.
Co-admitted states establish their baselines together, so only later blocks
contribute post-admission density growth. The backbone's native attention reads
the retained states directly from the resulting bank.
Figure~\ref{fig:method-overview} summarizes this online update.

Our contributions are:
\begin{itemize}
    \item We formulate historical KV maintenance as an online
    representation-coverage problem and derive an attention-aligned post-RoPE
    key geometry for
    query-agnostic redundancy estimation.
    \item We introduce DensityKV, combining per-head token banks, Soft-Riesz
    crowding, and insertion-relative density constraints while preserving exact
    K/V pairs without training, merging, or reconstruction.
    \item Across three backbones and 30--120-second generations, DensityKV
    achieves the best overall average rank and 31 of 54 single-metric wins under
    the same upper bound on historical KV capacity.
\end{itemize}

\section{Related Work}

\paragraph{Long video generation.}
One line of work extends pretrained bidirectional video diffusion models through
tuning-free noise rescheduling or FIFO denoising
\citep{qiu2023freenoise,kim2024fifo}. A complementary line distills short-clip
models into causal generators that produce clean frame blocks sequentially,
enabling low-latency rollouts beyond the training horizon
\citep{yin2024causvid,huang2025selfforcing,zhu2026causalforcing,
yang2025longlive}. These approaches improve generation schedules, causal
objectives, or rollout dynamics. Given a fixed AR backbone and local context,
DensityKV instead selects the completed historical K/V states available to
future blocks.

\paragraph{Memory for long video generation.}
Long-horizon video methods reintroduce discarded context in several ways.
Persistent sinks and structured temporal buffers retain early anchors or
selected historical frames \citep{yi2025deepforcing,zhao2026relaxforcing}.
Retrieval-based methods search generated history for non-local latent blocks,
dynamic anchors, trusted references, or scene memories
\citep{hu2026longliverag,ye2026dysink,meng2026tethercache,
wu2026echoforcing}. More recent token-level policies estimate salience from head
roles or future-query proxies, while latent-cache methods reduce the stored
representation itself \citep{chen2026pafukv,ji2026forcingkv,
luo2026futureforcing,yesiltepe2026videomla}. Most of these methods require a
retrieval signal, a predicted query, or a transformed cache representation.
DensityKV instead updates online as
each clean block becomes available, preserving exact K/V pairs using only native
post-RoPE key geometry. Unlike frame- or block-level selection, per-head token
banks let regions with different rates of change compete separately for
historical capacity.

\paragraph{Token-level KV bank management.}
KV-cache compression has been studied extensively for language models.
Attention-driven methods preserve sinks, heavy hitters, or tokens assigned high
head- and layer-specific importance
\citep{xiao2023streamingllm,zhang2023h2o,li2024snapkv,cai2024pyramidkv}.
Redundancy-aware methods further sample or cluster cached states, merge nearby
representatives, or organize semantic clusters for query-conditioned recall
\citep{zandieh2024subgen,cai2025rkv,hu2025centroidkv,liu2025clusterkv}.
Autoregressive video presents a distinct online setting: every generated block
adds dense spatiotemporal tokens, and repeatedly favoring the current query can
make the memory follow an already drifting trajectory. DensityKV therefore
maintains a query-agnostic subset independently in every attention head. It uses
post-RoPE Euclidean geometry and a Soft-Riesz crowding potential
\citep{hardin2005riesz}, while frozen insertion baselines distinguish newly
admitted support from density growth caused by later blocks.

\section{Method}

\subsection{Problem Setup}

Consider a frozen autoregressive video diffusion Transformer with $L$ layers,
$H$ attention heads, and a local K/V window. At update $t$, a finalized,
fully denoised block leaving the window supplies token-level candidates. Each
candidate uses its native per-head post-RoPE key as both attention key and
density descriptor and preserves its paired value. DensityKV retains at most
$M$ exact historical pairs per layer and head under an unbounded rollout and
unknown future queries. Figure~\ref{fig:method-overview} outlines the geometry,
density, and online update. Bold lowercase and calligraphic uppercase symbols
denote vectors and collections, respectively.

\subsection{Post-RoPE Attention Geometry}

Let $\mathbf q,\mathbf k\in\mathbb R^{d_k}$ denote the actual per-head query and
key vectors after the backbone applies RoPE. Their pre-softmax attention logit
is $\lambda(\mathbf q,\mathbf k)=\mathbf q^\top\mathbf k/\sqrt{d_k}$.
Cauchy--Schwarz yields
\begin{equation}
 \sup_{\mathbf q\neq\mathbf 0}
 \frac{\left|\lambda(\mathbf q,\mathbf k_i)
 -\lambda(\mathbf q,\mathbf k_j)\right|}{\lVert\mathbf q\rVert_2}
 =\frac{\lVert\mathbf k_i-\mathbf k_j\rVert_2}{\sqrt{d_k}}.
 \label{eq:logit-bound}
\end{equation}
Equality holds when $\mathbf q\parallel(\mathbf k_i-\mathbf k_j)$, so
Equation~\eqref{eq:logit-bound} gives the exact worst-case normalized logit
discrepancy, including native RoPE. We use squared Euclidean distance because
it preserves this neighborhood ordering, retains key-norm information absent
from cosine distance, and supports batched evaluation. This geometry motivates
a query-agnostic routing-coverage measure, not an attention-output error bound:
nearby keys need not have interchangeable values. DensityKV reallocates finite
capacity across key neighborhoods while copying or removing each selected K/V
pair exactly. Derivations are provided in the supplementary material.

\subsection{Soft-Riesz Density}

Classical Riesz energy uses inverse-power interactions to penalize crowded
point configurations \citep{hardin2005riesz}. We regularize this interaction
to obtain a finite, scale-aware kernel over the post-RoPE key geometry:
\begin{equation}
 w(\mathbf k_i,\mathbf k_j)=
 \left(\epsilon+
 \frac{\lVert\mathbf k_i-\mathbf k_j\rVert_2^2}{\sigma^2}\right)^{-p},
 \label{eq:riesz}
\end{equation}
where $\sigma$ sets neighborhood width, $p$ controls decay, and $\epsilon$
keeps the interaction finite when two keys coincide. Since $\epsilon^{-p}$ is
a common density scale and the effective bandwidth is
$\sigma\sqrt{\epsilon}$, we fix $\epsilon=1$ and use $(\sigma,p)=(8,2)$ by
default. For an indexed key multiset
$\mathcal A=(\mathbf k_j)_{j=1}^{|\mathcal A|}$, we define the
\emph{Soft-Riesz density} of entry $i$ as
\begin{equation}
 \Phi_{\mathcal A}(i)=
 \sum_{\substack{j=1\\j\neq i}}^{|\mathcal A|}
 w(\mathbf k_i,\mathbf k_j).
 \label{eq:density}
\end{equation}
High $\Phi_{\mathcal A}(i)$ indicates that many nearby keys already occupy the
bank. DensityKV
controls density growth \emph{after admission}, rather than absolute density.
The supplementary material gives the energy interpretation and calibration.

\subsection{Online Density-Constrained Bank Update}

\paragraph{Bank state and local baseline.}
At update $t$, layer $\ell$ and head $h$ maintain
\begin{equation}
 \mathcal{B}_{t}^{\ell,h}
 =\{(\bar{\mathbf k}_i^{\ell,h},
      \bar{\mathbf v}_i^{\ell,h},
      b_i^{\ell,h})\}_{i=1}^{m_t^\ell},
 \qquad m_t^\ell\leq M,
 \label{eq:bank}
\end{equation}
where $b_i^{\ell,h}$ is the insertion-density baseline and $m_t^\ell$ is the
occupancy shared across heads through synchronized admission. The $N$ arriving
states form
$\mathcal{C}_{t}^{\ell,h}
 =\{(\mathbf k_x^{\ell,h},\mathbf v_x^{\ell,h})\}_{x=1}^{N}$.
For any K/V collection $\mathcal S$, let $\mathcal K(\mathcal S)$ denote its
indexed post-RoPE key multiset, ignoring auxiliary fields such as $b_i$.
Separate banks reflect head-specific projection geometries. We suppress $t$ and
$\ell$ when unambiguous.

Each admitted historical KV state
$(\bar{\mathbf k}_i,\bar{\mathbf v}_i)$ stores a frozen insertion-density
baseline for its post-RoPE key,
\begin{equation}
 b_i=\max\left(
 \Phi_{\mathcal K_i^{\mathrm{ins}}}(i),\delta
 \right),
 \label{eq:insertion-baseline}
\end{equation}
where $\mathcal K_i^{\mathrm{ins}}$ is the final key multiset produced by the
update that admits state $i$, and $\delta>0$ is a numerical floor. The frozen
baseline measures subsequent growth relative to the neighborhood at admission.

\begin{figure*}[!t]
\centering
\includegraphics[width=0.96\textwidth]{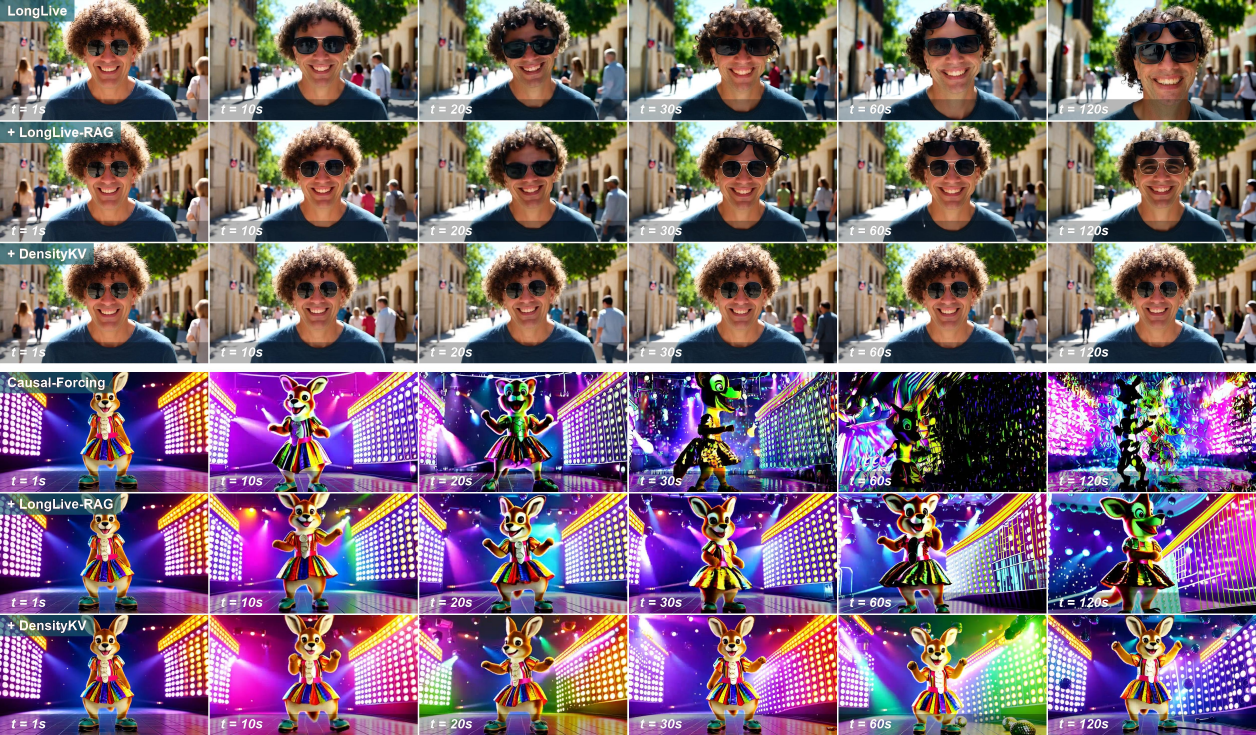}
\caption{Qualitative comparison on two matched backbone--prompt pairs, each
generated for 120 seconds. The upper group uses LongLive on MovieGenBench
\#097, the wig-and-sunglasses transformation; the lower group uses Causal-Forcing on
MovieGenBench \#012, the disco-dancing kangaroo. }
\label{fig:longlive-qualitative}
\vspace{-10pt} 
\end{figure*}

\ifvbenchresults
\begin{figure*}[!t]
\centering
\begin{minipage}{0.98\textwidth}
\centering
\scriptsize
\setlength{\tabcolsep}{2.7pt}
\renewcommand{\arraystretch}{0.88}
\begin{tabular}{l@{\hspace{0.85em}}llrrrrrrr}
\toprule
Time & Backbone & Method & Subject$\uparrow$ & Background$\uparrow$ &
Smoothness$\uparrow$ & Dynamic$\uparrow$ & Aesthetic$\uparrow$ &
Imaging$\uparrow$ & Avg. Rank$\downarrow$ \\
\midrule
\input{generated/full_vbench_30_60_120_rows.tex}\\
\addlinespace[0.8pt]
\bottomrule
\end{tabular}
\captionof{table}{Comparison across backbones. Bold/underline mark best/second-best values; Avg. Rank is averaged over the six metrics.}
\label{tab:vbench-main}
\end{minipage}
\vspace{-10pt} 
\end{figure*}
\fi

\ifdensitybthreeablations
\ifdensitybthreemechanisms
\ifdensitybthreeheadwise
\begin{table*}[t]
\centering
\fontsize{7.2}{8.0}\selectfont
\setlength{\tabcolsep}{2.8pt}
\renewcommand{\arraystretch}{0.88}
\begin{tabular*}{\textwidth}{@{\extracolsep{\fill}}llrrrrrrr}
\toprule
\multicolumn{9}{c}{\textbf{(a) Parameter sensitivity}}\\
Parameter & Setting & Subject$\uparrow$ & Background$\uparrow$ &
Smoothness$\uparrow$ & Dynamic$\uparrow$ &
Aesthetic$\uparrow$ & Imaging$\uparrow$ & Avg. Rank$\downarrow$\\
\midrule
% Generated by scripts/render_densitykv_b3_crossbackbone_ablations.py.
% 60-second metrics are macro means over three backbones.
% Starred rows use the mean-normalized default policy.
% Avg. Rank averages ranks over the six displayed macro metrics.
\multirow{3}{*}{$\tau$} & $1.5$ & 97.03 & 96.29 & 98.26 & \textbf{52.57} & \textbf{57.97} & \underline{68.70} & 2.17 \\
 & $2.0^{*}$ & \underline{97.19} & \textbf{96.34} & \underline{98.33} & \underline{51.94} & 57.74 & \textbf{68.73} & \textbf{1.83} \\
 & $3.0$ & \textbf{97.20} & \underline{96.33} & \textbf{98.35} & 44.44 & \underline{57.83} & 65.50 & \underline{2.00} \\
\midrule
\multirow{3}{*}{$M$} & 6{,}240 & \textbf{97.25} & \underline{96.31} & \textbf{98.34} & 49.38 & 57.61 & 67.06 & 2.17 \\
 & 9{,}360$^{*}$ & \underline{97.19} & \textbf{96.34} & \underline{98.33} & \textbf{51.94} & \underline{57.74} & \underline{68.73} & \textbf{1.67} \\
 & 12{,}480 & 97.12 & 96.30 & \underline{98.33} & \underline{51.39} & \textbf{57.85} & \textbf{69.21} & \underline{2.00} \\
\midrule
\multirow{3}{*}{$\sigma$} & $4$ & 97.08 & \underline{96.28} & 98.28 & \textbf{52.64} & \underline{57.78} & \underline{68.85} & 2.17 \\
 & $8^{*}$ & \textbf{97.19} & \textbf{96.34} & \textbf{98.33} & \underline{51.94} & 57.74 & 68.73 & \textbf{1.83} \\
 & $16$ & \underline{97.18} & 96.27 & \underline{98.32} & 50.49 & \textbf{57.94} & \textbf{69.27} & \underline{2.00} \\
\midrule
\multirow{3}{*}{$p$} & $1$ & \textbf{97.19} & \underline{96.32} & \underline{98.32} & 50.35 & \underline{57.71} & \textbf{69.02} & \underline{1.83} \\
 & $2^{*}$ & \textbf{97.19} & \textbf{96.34} & \textbf{98.33} & \textbf{51.94} & \textbf{57.74} & \underline{68.73} & \textbf{1.17} \\
 & $4$ & \underline{97.06} & 96.27 & 98.22 & \underline{51.88} & 57.68 & 68.52 & 2.83
\\
\addlinespace[1pt]
\midrule
\multicolumn{9}{c}{\textbf{(b) Policy ablations}}\\
\midrule
% Generated by scripts/render_densitykv_b3_crossbackbone_ablations.py.
% 60-second metrics are macro means over three backbones.
% Bold/underline are computed within each policy group.
Default$^{*}$ & & \textbf{97.19} & \textbf{96.34} & 98.33 & \underline{51.94} & 57.74 & 68.73 & \textbf{2.83} \\
Refresh density baselines & & \textbf{97.19} & 96.30 & \textbf{98.42} & 47.29 & 56.84 & 64.85 & 4.33 \\
Source-order candidates & & 97.06 & 96.30 & 98.28 & 51.32 & 57.36 & 68.54 & 5.17 \\
Densest-only eviction & & 97.12 & 96.28 & \underline{98.34} & 51.53 & 57.85 & \underline{69.22} & \underline{3.17} \\
Mandatory + source eviction & & 97.12 & 96.28 & \underline{98.34} & 51.11 & \underline{57.86} & \textbf{69.23} & \underline{3.17} \\
Independent per-head & & \underline{97.14} & \underline{96.31} & 98.30 & 50.00 & \textbf{58.14} & 69.05 & 3.33 \\
Within-block competition & & 96.96 & 96.16 & 98.10 & \textbf{57.50} & 57.69 & 68.83 & 5.17 \\
\midrule
\multirow{4}{*}{Geometry} & pre-RoPE $K$ & \underline{96.93} & 96.18 & \underline{98.23} & \underline{51.88} & \textbf{57.38} & \textbf{67.79} & 2.33 \\
 & pre-RoPE $[K;V]$ & \underline{96.93} & \underline{96.22} & \textbf{98.28} & \textbf{52.64} & 57.24 & 66.41 & \underline{2.00} \\
 & post-RoPE $K$ & \textbf{97.01} & \textbf{96.23} & \textbf{98.28} & 49.65 & \underline{57.31} & \underline{67.01} & \textbf{1.83} \\
 & post-RoPE $[K;V]$ & 96.88 & 96.20 & \textbf{98.28} & 50.56 & 57.17 & 65.91 & 3.17
\\
\addlinespace[1pt]
\bottomrule
\end{tabular*}
\caption{Parameter sensitivity and policy ablations. Bold/underline mark
best/second-best displayed values within each ablation group.}
\label{tab:densitykv-ablations}
\vspace{-10pt} 
\end{table*}
\fi
\fi
\fi

\paragraph{Step 1: rank candidates by mean-normalized disturbance.}
For a nonempty bank, candidate $x$ in head $h$ receives
\begin{equation}
 s_h(x)=
 \frac{1}{m_t}
 \sum_{i=1}^{m_t}
 \frac{w(\bar{\mathbf k}_i^h,\mathbf k_x^h)}{b_i^h},
 \label{eq:candidate-score}
\end{equation}
which averages the normalized one-candidate increment over the bank. Each head
sorts candidates by increasing $s_h$, producing permutation $\pi_h$ and prefix
\begin{equation}
 \mathcal P_r^h
 =
 \left\{
 (\mathbf k_{\pi_h(j)}^h,\mathbf v_{\pi_h(j)}^h)
 \right\}_{j=1}^{r}.
 \label{eq:candidate-prefix}
\end{equation}
We set $\mathcal P_0^h=\varnothing$. This step only determines the ordering;
the next step evaluates cumulative prefix effects before admission.

\paragraph{Step 2: find the largest feasible cross-head prefix.}
If $\mathcal P_r^h$ were admitted, the projected density-growth ratio of
retained state $i$ would be
\begin{equation}
 r_i^h(\mathcal B^h,\mathcal P_r^h)=
 \frac{
 \Phi_{\mathcal K(\mathcal B^h)}(i)+
 \sum_{j=1}^{r}
 w\!\left(\bar{\mathbf k}_i^h,\mathbf k_{\pi_h(j)}^h\right)}
 {b_i^h}.
 \label{eq:density-growth}
\end{equation}
The numerator combines current density with the prefix's cumulative
contribution. Given a maximum growth factor $\tau>1$, let
\begin{equation}
 v_r^h=
 \left|
 \left\{
 i:r_i^h(\mathcal B^h,\mathcal P_r^h)\geq\tau
 \right\}
 \right|
 \label{eq:violator-count}
\end{equation}
count the historical states whose projected ratio reaches $\tau$. Separately,
admitting $r$ candidates requires
\begin{equation}
 e_r=\max(0,m_t+r-M)
 \label{eq:eviction-budget}
\end{equation}
capacity-driven evictions. A prefix is feasible when $v_r^h\leq e_r$, so all
violating historical states fit within the required eviction budget. The layer
selects
\begin{equation}
 r^\star=
 \max_{\substack{0\leq r\leq\min(N,M)\\
                   v_r^h\leq e_r,\ \forall h}} r,
 \qquad r\in\mathbb Z.
 \label{eq:shared-prefix}
\end{equation}
All stored states satisfy the density-growth bound before an update, which makes
$r=0$ feasible. The most restrictive head therefore controls the shared
cardinality; each head still uses its own
$\mathcal P_{r^\star}^h$. The bank may remain below $M$ when no positive prefix
is feasible.

\paragraph{Step 3: evict, commit, and preserve the invariant.}
For each head, all retained states satisfying
$r_i^h(\mathcal B^h,\mathcal P_{r^\star}^h)\geq\tau$ are marked for mandatory
eviction.
If there are fewer than $e_{r^\star}$ such states, DensityKV removes additional
historical states in decreasing projected Soft-Riesz density until exactly
$e_{r^\star}$ slots are released. With eviction set $\mathcal R^h$, it commits
\begin{equation}
 \mathcal B_{t+1}^h
 =
 \left(\mathcal B_t^h\setminus\mathcal R^h\right)
 \cup\mathcal P_{r^\star}^h.
 \label{eq:bank-commit}
\end{equation}
Since $w\geq0$, evicting states cannot increase survivor density, so all
surviving historical states satisfy the growth bound. New states derive $b_i$
from the final bank: same-block candidate interactions become part of their
frozen admission baselines, whereas later blocks contribute post-admission
growth. Thus, the feasibility test constrains the cumulative effect of an
admitted prefix on existing history, whereas the admitted candidates establish
their baselines jointly. Keys determine all decisions, and values move unchanged
with their paired keys. An empty bank
admits up to $M$ source-ordered candidates and computes their baselines in the
resulting bank. Pseudocode is provided in the supplementary material.

For $T$ finalized tokens per head, full history requires
$O(LHT(d_k+d_v))$ persistent K/V storage; DensityKV requires
$O(LHM(d_k+d_v))$, independent of rollout length.

\section{Experiments}
\label{sec:experiments}

\subsection{Implementation Details}

We build on the open Wan2.1-T2V-1.3B model \citep{wan2025} and use the
released frozen checkpoints for LongLive, Self-Forcing, and Causal-Forcing.
DensityKV requires no model fine-tuning. For quality comparisons, we match
methods by the upper bound on attention-visible nonlocal K/V tokens rather than
by instantaneous occupancy. Each method receives a nonlocal budget equivalent
to six complete frames per head, in addition to one explicit first-frame sink
and a five-frame local window. DensityKV therefore stores at most 9,360
historical KV states per attention head. Unless stated otherwise, we use
$\epsilon=1$, $p=2$, $\sigma=8$, $\delta=10^{-6}$, and density-growth
threshold $\tau=2.0$. The bank is reset for each video.

\begin{figure}[t]
    \centering
    \includegraphics[width=0.88\columnwidth]{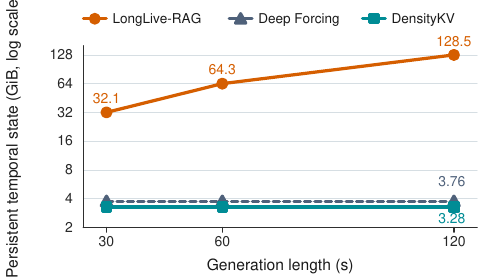}
    \caption{Persistent temporal-state storage versus generation length.
    LongLive-RAG grows linearly, whereas Deep Forcing and DensityKV remain
    bounded.}
    \label{fig:storage-scaling}
\end{figure}

\paragraph{Persistent temporal-state storage.}
We compare the aggregate persistent temporal state retained on CPU and GPU,
excluding model parameters and layer-local temporary workspace for every
method. LongLive-RAG retains the raw BF16 K/V states of all latent frames
together with their retrieval descriptors, requiring 32.1, 64.3, and
128.5 GiB at 30, 60, and 120 seconds, respectively. Deep Forcing retains a
12-frame K/V cache, a four-frame query buffer, and metadata, totaling 3.76 GiB.
At $M=9{,}360$, DensityKV's post-RoPE K/V bank, density baselines, and
provenance occupy 1.67 GiB; including its first-frame sink and five-frame local
K/V window yields 3.28 GiB of total persistent temporal-state storage. DensityKV thus
reduces storage relative to LongLive-RAG by $9.8{\times}$, $19.6{\times}$, and
$39.2{\times}$ at the same horizons (Figure~\ref{fig:storage-scaling}).

\subsection{General Setup}

Our main evaluation follows LongLive-RAG's long-video protocol
\citep{hu2026longliverag}, using all 128 MovieGenBench prompts refined with
Qwen2.5-7B-Instruct. We compare Native, $\infty$-RoPE, Deep Forcing,
LongLive-RAG, and DensityKV on LongLive, Self-Forcing, and Causal-Forcing.
We use the baseline scores reported by LongLive-RAG under this protocol and
evaluate DensityKV with matched released checkpoints, prompts, sampler settings,
and seeds.

Controlled ablations use a fixed 16-prompt subset from the same benchmark.
Parameter sensitivity sweeps $\tau$, bank capacity $M$, $\sigma$, and $p$
around the default
$(2.0,9{,}360,8,2)$, with $\epsilon=1$ fixed by the scale normalization above;
discrete ablations change exactly one policy component.

\subsection{Metrics}

VBench defines 16 dimensions \citep{huang2023vbench}; following LongLive-RAG,
we report the six used by its long-video protocol: Subject Consistency,
Background Consistency, Motion Smoothness, Dynamic Degree, Aesthetic Quality,
and Imaging Quality. Higher is better for all six. Avg. Rank averages method
ranks across these metrics (lower is better); three-backbone ablations report
unweighted means across backbones.

\subsection{Main Results}
\noindent\textbf{Comparison across backbones.}

Figure~\ref{fig:longlive-qualitative} compares the native backbone,
LongLive-RAG, and DensityKV on two matched cases. DensityKV better preserves
identity, hairstyle, sunglasses, and street structure on LongLive. On
Causal-Forcing, the native model develops late subject and scene distortions;
LongLive-RAG mitigates them, while DensityKV preserves subject appearance at
60 seconds. These visual comparisons are consistent with the quantitative
results below.

\ifvbenchresults
\begin{figure}[t]
\centering
\includegraphics[width=0.95\linewidth]{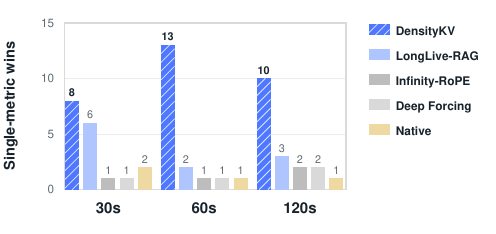}
\caption{Single-metric wins under the matched LongLive-RAG protocol
at 30, 60, and 120 seconds.}
\label{fig:vbench-wins}
\end{figure}

Table~\ref{tab:vbench-main} compares five methods on three backbones, three
horizons, and the six selected VBench metrics. Figure~\ref{fig:vbench-wins} summarizes
single-metric wins: DensityKV leads at every horizon with 8, 13, and 10 wins at
30, 60, and 120 seconds. Overall, it obtains 31 of 54 wins and the best Avg.
Rank of 1.89, while improving Background Consistency over LongLive-RAG in all
nine horizon--backbone comparisons.

Gains vary by backbone. DensityKV ranks first at all Self-Forcing horizons. On
Causal-Forcing it reaches an Avg. Rank of 1.33 at 60 and
120 seconds and achieves the best scores on five metrics, while $\infty$-RoPE achieves a higher
Dynamic Degree. This trade-off suggests that stronger historical anchoring can reduce
motion diversity on this high-motion backbone. On LongLive, DensityKV achieves
the best Background Consistency, Dynamic Degree, and Imaging Quality at 60 and
120 seconds, whereas LongLive-RAG retains stronger Subject Consistency and
Aesthetic Quality.
Overall, DensityKV transfers across all three backbones, with gains shaped by
each generator's error profile.

\fi

\subsection{Ablation Study}

\ifdensitybthreeablations
\paragraph{Parameter sensitivity.}
Table~\ref{tab:densitykv-ablations}(a) varies the density-growth threshold,
capacity, and kernel shape around
$(\tau,M,\sigma,p)=(2.0,9{,}360,8,2)$ with $\epsilon=1$ fixed. Avg.\ Rank is computed over
the six displayed macro-averaged metrics within each sweep. At 60 seconds,
$\tau=2.0$ ranks best at 1.83, while $\tau=1.5$ favors dynamics and aesthetics
and $\tau=3.0$ favors consistency. Performance varies non-monotonically with capacity:
$M=9{,}360$ ranks best at 1.67, balancing the stronger subject consistency of
$M=6{,}240$ with the visual quality of $M=12{,}480$. Increasing $\sigma$ to 16
improves aesthetic and imaging quality, whereas $\sigma=8$ improves subject,
background, smoothness, and dynamics and obtains the best Avg.\ Rank of 1.83.
The default $p=2$ ranks best in its sweep at 1.17. We therefore use
$(2.0,9{,}360,8,2)$ as the balanced default across consistency, dynamics, and
visual quality, with $\epsilon=1$ serving only as the fixed kernel normalization.
\fi

\ifdensitybthreemechanisms
\paragraph{Policy ablations.}
Table~\ref{tab:densitykv-ablations}(b) isolates candidate ordering, density
baselines, eviction selection, cross-head synchronization, within-block
competition, and retention geometry.

Candidate ordering has the largest effect among the tested update policies.
Removing density guidance and processing candidates in source order gives an
Avg.\ Rank of 5.17, compared with 2.83 for the density-guided default. The
mean-normalized score prevents one sensitive historical state from determining
the candidate order and favors candidates with a small average disturbance
across the bank.

Freezing the density baseline recorded at admission is also important.
Refreshing the baselines after every update attains the highest Motion
Smoothness of 98.42, but reduces Imaging Quality to 64.85 and gives an Avg.\ Rank
of 4.33. With continuous baseline updates, later density growth is measured
against an already crowded bank, which weakens the density-growth constraint
tied to admission.

The eviction variants are more stable than the admission variants.
Densest-only eviction and source-ordered auxiliary eviction both obtain an
Avg.\ Rank of 3.17, but the former better preserves dynamics while the latter
favors aesthetics and imaging quality. This indicates that eviction selection
primarily controls the trade-off among these criteria.

Allowing each attention head to choose its admission count independently
raises Aesthetic Quality from 57.74 to 58.14 and Imaging Quality from 68.73 to
69.05, but slightly lowers Subject Consistency, Background Consistency, Motion
Smoothness, and Dynamic Degree, yielding an Avg.\ Rank of 3.33. Sharing the
admission count keeps the number of admitted states aligned across heads and
gives a better overall rank.

The retention-geometry ablation supports measuring redundancy with post-RoPE
keys alone. Post-RoPE $K$ obtains the best Avg.\ Rank of 1.83, achieving the
highest Subject and Background Consistency and tying for the best Motion
Smoothness. Its lower
Dynamic Degree (49.65) reflects a stability--dynamics trade-off: stronger
attention-aligned historical anchoring improves long-horizon consistency while
reducing motion diversity. Pre-RoPE $K$ and $[K;V]$ obtain Avg.\ Ranks of 2.33
and 2.00, respectively, while post-RoPE $[K;V]$ obtains the worst Avg.\ Rank of
3.17. This result is
consistent with the role of keys in attention: queries address historical
states through post-RoPE key similarity, whereas values determine the returned
content but do not define the query-to-memory correspondence. We therefore use
post-RoPE $K$ alone for all density decisions.

Immediate within-block competition evaluates insertion hysteresis by
performing an additional density cleanup after committing each candidate block,
so co-admitted states compete before their baselines are frozen. It raises
Dynamic Degree to 57.50, but gives the lowest Subject Consistency, Background
Consistency, and Motion Smoothness and an Avg.\ Rank of 5.17. This result
suggests that immediate absolute-density balancing can remove coherent
within-block evidence; delaying this competition preserves local continuity,
while subsequent blocks still trigger post-admission density-growth eviction.

Overall, density-guided candidate ordering and frozen insertion baselines are
the most consequential policy components. Eviction and cross-head
synchronization primarily control the trade-off among consistency, dynamics,
and visual quality. The default obtains the best Avg.\ Rank among update-policy
variants (2.83) and is therefore used in the main experiments.
\fi

\section{Conclusion}

DensityKV is a training-free historical KV selection algorithm that controls
post-admission crowding in post-RoPE key space while preserving exact K/V
pairs. Across three frozen video backbones, it improves long-horizon consistency
under the same historical-capacity bound and keeps storage independent of
rollout length. By allocating history at token and head granularity, the method
preserves non-local subject and scene evidence without training or modifying the
generator. Because retained values are neither merged nor reconstructed, every
selected state remains directly usable by the backbone's native attention.

\clearpage
\bibliography{references}

\begin{thebibliography}{29}
\providecommand{\natexlab}[1]{#1}

\bibitem[{Cai et~al.(2025)Cai, Xiao, Sun, Luo, Zhang, Wan, Li, Zhou, Chang, Gu,
  Dong, Anandkumar, Asi, and Hu}]{cai2025rkv}
Cai, Z.; Xiao, W.; Sun, H.; Luo, C.; Zhang, Y.; Wan, K.; Li, Y.; Zhou, Y.;
  Chang, L.-W.; Gu, J.; Dong, Z.; Anandkumar, A.; Asi, A.; and Hu, J. 2025.
\newblock R-KV: Redundancy-aware KV Cache Compression for Reasoning Models.
\newblock In \emph{Advances in Neural Information Processing Systems}.

\bibitem[{Cai et~al.(2024)Cai, Zhang, Gao, Liu, Liu, Lu, Xiong, Dong, Chang,
  Hu, and Xiao}]{cai2024pyramidkv}
Cai, Z.; Zhang, Y.; Gao, B.; Liu, Y.; Liu, T.; Lu, K.; Xiong, W.; Dong, Y.;
  Chang, B.; Hu, J.; and Xiao, W. 2024.
\newblock PyramidKV: Dynamic KV Cache Compression based on Pyramidal
  Information Funneling.
\newblock \emph{arXiv preprint arXiv:2406.02069}.

\bibitem[{Chen et~al.(2026)Chen, Xu, Yang, Chen, and Deng}]{chen2026pafukv}
Chen, H.; Xu, C.; Yang, X.; Chen, X.; and Deng, C. 2026.
\newblock Past- and Future-Informed KV Cache Policy with Salience Estimation in
  Autoregressive Video Diffusion.
\newblock \emph{arXiv preprint arXiv:2601.21896}.

\bibitem[{Hardin and Saff(2005)}]{hardin2005riesz}
Hardin, D.~P.; and Saff, E.~B. 2005.
\newblock Minimal Riesz Energy Point Configurations for Rectifiable
  $d$-Dimensional Manifolds.
\newblock \emph{Advances in Mathematics}, 193(1): 174--204.

\bibitem[{Hu et~al.(2025)Hu, Wang, He, Gong, Yi, Zhang, Bai, Chen, Zhang, Li,
  and Yuan}]{hu2025centroidkv}
Hu, J.; Wang, S.; He, Y.; Gong, P.; Yi, J.; Zhang, J.; Bai, Y.; Chen, R.;
  Zhang, G.; Li, C.; and Yuan, K. 2025.
\newblock CentroidKV: Efficient Long-Context LLM Inference via KV Cache
  Clustering.
\newblock \emph{arXiv preprint arXiv:2506.11418}.

\bibitem[{Hu et~al.(2026)Hu, Yang, Huang, Han, and Chen}]{hu2026longliverag}
Hu, Q.; Yang, S.; Huang, W.; Han, S.; and Chen, Y. 2026.
\newblock LongLive-RAG: A General Retrieval-Augmented Framework for Long Video
  Generation.
\newblock \emph{arXiv preprint arXiv:2606.02553}.

\bibitem[{Huang et~al.(2025)Huang, Li, He, Zhou, and
  Shechtman}]{huang2025selfforcing}
Huang, X.; Li, Z.; He, G.; Zhou, M.; and Shechtman, E. 2025.
\newblock Self Forcing: Bridging the Train-Test Gap in Autoregressive Video
  Diffusion.
\newblock \emph{arXiv preprint arXiv:2506.08009}.

\bibitem[{Huang et~al.(2023)Huang, He, Yu, Zhang, Si, Jiang, Zhang, Wu, Jin,
  Chanpaisit et~al.}]{huang2023vbench}
Huang, Z.; He, Y.; Yu, J.; Zhang, F.; Si, C.; Jiang, Y.; Zhang, Y.; Wu, T.;
  Jin, Q.; Chanpaisit, N.; et~al. 2023.
\newblock VBench: Comprehensive Benchmark Suite for Video Generative Models.
\newblock \emph{arXiv preprint arXiv:2311.17982}.

\bibitem[{Ji et~al.(2026)Ji, Zhong, Zhang, Yang, Jin, Qin, Luo, Mao, Liu, and
  Li}]{ji2026forcingkv}
Ji, Y.; Zhong, Z.; Zhang, J.; Yang, Q.; Jin, X.; Qin, Y.; Luo, W.; Mao, S.;
  Liu, W.; and Li, H. 2026.
\newblock Forcing-KV: Hybrid KV Cache Compression for Efficient Autoregressive
  Video Diffusion Models.
\newblock \emph{arXiv preprint arXiv:2605.09681}.

\bibitem[{Kim et~al.(2024)Kim, Kang, Choi, and Han}]{kim2024fifo}
Kim, J.; Kang, J.; Choi, J.; and Han, B. 2024.
\newblock FIFO-Diffusion: Generating Infinite Videos from Text without
  Training.
\newblock \emph{arXiv preprint arXiv:2405.11473}.

\bibitem[{Li et~al.(2024)Li, Huang, Yang, Venkitesh, Locatelli, Ye, Cai, Lewis,
  and Chen}]{li2024snapkv}
Li, Y.; Huang, Y.; Yang, B.; Venkitesh, B.; Locatelli, A.; Ye, H.; Cai, T.;
  Lewis, P.; and Chen, D. 2024.
\newblock SnapKV: LLM Knows What You Are Looking for Before Generation.
\newblock \emph{Advances in Neural Information Processing Systems}, 37.

\bibitem[{Liu et~al.(2025)Liu, Li, Zhao, Zhang, and Guo}]{liu2025clusterkv}
Liu, G.; Li, C.; Zhao, J.; Zhang, C.; and Guo, M. 2025.
\newblock ClusterKV: Manipulating LLM KV Cache in Semantic Space for Recallable
  Compression.
\newblock In \emph{Proceedings of the 62nd ACM/IEEE Design Automation
  Conference}, 1--7.

\bibitem[{Luo et~al.(2026)Luo, Liu, Wang, Liu, Chen, Wang, Zhu, Gao, Hu, Sun,
  and Chen}]{luo2026futureforcing}
Luo, J.; Liu, Q.; Wang, T.; Liu, J.; Chen, J.; Wang, C.; Zhu, H.; Gao, C.; Hu,
  X.; Sun, Q.; and Chen, Z. 2026.
\newblock Future Forcing: Future-aware Training-free KV Cache Policy for
  Autoregressive Video Generation.
\newblock \emph{arXiv preprint arXiv:2605.30083}.

\bibitem[{Meng et~al.(2026)Meng, Luo, Li, Jiang, Gao, Chen, Li, and
  Zhang}]{meng2026tethercache}
Meng, Y.; Luo, X.; Li, L.; Jiang, W.; Gao, C.; Chen, X.; Li, Y.; and Zhang,
  X.-P. 2026.
\newblock TetherCache: Stabilizing Autoregressive Long-Form Video Generation
  with Gated Recall and Trusted Alignment.
\newblock \emph{arXiv preprint arXiv:2606.13035}.

\bibitem[{Peebles and Xie(2023)}]{peebles2023dit}
Peebles, W.; and Xie, S. 2023.
\newblock Scalable Diffusion Models with Transformers.
\newblock In \emph{Proceedings of the IEEE/CVF International Conference on
  Computer Vision}, 4195--4205.

\bibitem[{Qiu et~al.(2023)Qiu, Xia, Zhang, He, Wang, Shan, and
  Liu}]{qiu2023freenoise}
Qiu, H.; Xia, M.; Zhang, Y.; He, Y.; Wang, X.; Shan, Y.; and Liu, Z. 2023.
\newblock FreeNoise: Tuning-Free Longer Video Diffusion via Noise Rescheduling.
\newblock \emph{arXiv preprint arXiv:2310.15169}.

\bibitem[{{Wan Team} et~al.(2025){Wan Team}, Wang, Ai, Wen, Mao, Xie, Chen, Yu,
  Zhao, Yang et~al.}]{wan2025}
{Wan Team}; Wang, A.; Ai, B.; Wen, B.; Mao, C.; Xie, C.-W.; Chen, D.; Yu, F.;
  Zhao, H.; Yang, J.; et~al. 2025.
\newblock Wan: Open and Advanced Large-Scale Video Generative Models.
\newblock \emph{arXiv preprint arXiv:2503.20314}.

\bibitem[{Wu et~al.(2026)Wu, Feng, Zhang, Qin, Li, Fan, Liu, An, Huang, Xu, and
  Yang}]{wu2026echoforcing}
Wu, M.; Feng, W.; Zhang, Z.; Qin, H.; Li, Y.; Fan, G.; Liu, X.; An, Z.; Huang,
  L.; Xu, Y.; and Yang, C. 2026.
\newblock Echo-Forcing: A Scene Memory Framework for Interactive Long Video
  Generation.
\newblock \emph{arXiv preprint arXiv:2605.16003}.

\bibitem[{Xiao et~al.(2023)Xiao, Tian, Chen, Han, and
  Lewis}]{xiao2023streamingllm}
Xiao, G.; Tian, Y.; Chen, B.; Han, S.; and Lewis, M. 2023.
\newblock Efficient Streaming Language Models with Attention Sinks.
\newblock \emph{arXiv preprint arXiv:2309.17453}.

\bibitem[{Yang et~al.(2025)Yang, Huang, Chu, Xiao, Zhao, Wang, Li, Xie, Chen,
  Lu, Han, and Chen}]{yang2025longlive}
Yang, S.; Huang, W.; Chu, R.; Xiao, Y.; Zhao, Y.; Wang, X.; Li, M.; Xie, E.;
  Chen, Y.; Lu, Y.; Han, S.; and Chen, Y. 2025.
\newblock LongLive: Real-time Interactive Long Video Generation.
\newblock \emph{arXiv preprint arXiv:2509.22622}.

\bibitem[{Ye et~al.(2026)Ye, Cui, Zhao, Wei, and Zhang}]{ye2026dysink}
Ye, B.; Cui, X.; Zhao, J.; Wei, T.; and Zhang, M.-L. 2026.
\newblock DySink: Dynamic Frame Sinks for Autoregressive Long Video Generation.
\newblock \emph{arXiv preprint arXiv:2605.21028}.

\bibitem[{Yesiltepe et~al.(2026)Yesiltepe, Hu, Meral, Akan, Oktay, Eldardiry,
  and Yanardag}]{yesiltepe2026videomla}
Yesiltepe, H.; Hu, J.; Meral, T. H.~S.; Akan, A.~K.; Oktay, K.; Eldardiry, H.;
  and Yanardag, P. 2026.
\newblock VideoMLA: Low-Rank Latent KV Cache for Minute-Scale Autoregressive
  Video Diffusion.
\newblock \emph{arXiv preprint arXiv:2605.30351}.

\bibitem[{Yesiltepe et~al.(2025)Yesiltepe, Meral, Akan, Oktay, and
  Yanardag}]{yesiltepe2025infinityrope}
Yesiltepe, H.; Meral, T. H.~S.; Akan, A.~K.; Oktay, K.; and Yanardag, P. 2025.
\newblock Infinity-RoPE: Action-Controllable Infinite Video Generation Emerges
  From Autoregressive Self-Rollout.
\newblock \emph{arXiv preprint arXiv:2511.20649}.

\bibitem[{Yi et~al.(2025)Yi, Jang, Cho, Nam, Yoon, and Kim}]{yi2025deepforcing}
Yi, J.; Jang, W.; Cho, P.~H.; Nam, J.; Yoon, H.; and Kim, S. 2025.
\newblock Deep Forcing: Training-Free Long Video Generation with Deep Sink and
  Participative Compression.
\newblock \emph{arXiv preprint arXiv:2512.05081}.

\bibitem[{Yin et~al.(2024)Yin, Zhang, Zhang, Freeman, Durand, Shechtman, and
  Huang}]{yin2024causvid}
Yin, T.; Zhang, Q.; Zhang, R.; Freeman, W.~T.; Durand, F.; Shechtman, E.; and
  Huang, X. 2024.
\newblock From Slow Bidirectional to Fast Autoregressive Video Diffusion
  Models.
\newblock \emph{arXiv preprint arXiv:2412.07772}.

\bibitem[{Zandieh et~al.(2024)Zandieh, Han, Mirrokni, and
  Karbasi}]{zandieh2024subgen}
Zandieh, A.; Han, I.; Mirrokni, V.; and Karbasi, A. 2024.
\newblock SubGen: Token Generation in Sublinear Time and Memory.
\newblock \emph{arXiv preprint arXiv:2402.06082}.

\bibitem[{Zhang et~al.(2023)Zhang, Sheng, Zhou, Chen, Zheng, Cai, Song, Tian,
  R\'{e}, Barrett, Wang, and Chen}]{zhang2023h2o}
Zhang, Z.; Sheng, Y.; Zhou, T.; Chen, T.; Zheng, L.; Cai, R.; Song, Z.; Tian,
  Y.; R\'{e}, C.; Barrett, C.; Wang, Z.; and Chen, B. 2023.
\newblock H$_2$O: Heavy-Hitter Oracle for Efficient Generative Inference of
  Large Language Models.
\newblock \emph{Advances in Neural Information Processing Systems}, 36.

\bibitem[{Zhao et~al.(2026)Zhao, Lu, Liu, Song, Deng, and
  Patras}]{zhao2026relaxforcing}
Zhao, Z.; Lu, Y.; Liu, Z.; Song, J.; Deng, J.; and Patras, I. 2026.
\newblock Relax Forcing: Relaxed KV-Memory for Consistent Long Video
  Generation.
\newblock \emph{arXiv preprint arXiv:2603.21366}.

\bibitem[{Zhu et~al.(2026)Zhu, Zhao, He, Su, Li, and
  Zhu}]{zhu2026causalforcing}
Zhu, H.; Zhao, M.; He, G.; Su, H.; Li, C.; and Zhu, J. 2026.
\newblock Causal Forcing: Autoregressive Diffusion Distillation Done Right for
  High-Quality Real-Time Interactive Video Generation.
\newblock \emph{arXiv preprint arXiv:2602.02214}.

\end{thebibliography}

\ifnum\includesupplementarypreview=1\relax
\FloatBarrier
\clearpage
\setcounter{section}{0}
\setcounter{subsection}{0}
\setcounter{equation}{0}
\renewcommand{\thesection}{\Alph{section}}
\section*{Supplementary Material}
% Shared supplementary content. This file is included both by paper.tex for the
% combined preprint and by supplementary-standalone.tex for optional standalone
% distribution. Method definitions follow the main paper.
\raggedbottom
\setcounter{secnumdepth}{2}
\numberwithin{equation}{section}

\providecommand{\suppcausalcasefigure}{figures/supplementary_figure3_extension_a.pdf}
\providecommand{\suppselfcasefigure}{figures/supplementary_figure3_extension_b.pdf}
\providecommand{\supplonglivecasefigure}{figures/supplementary_figure3_extension_c.pdf}

\paragraph{Organization.}
Appendices~\ref{sec:supp-distance}--\ref{sec:supp-update-algorithm} derive the
post-RoPE key geometry and Soft-Riesz density, then give the complete online
bank update. Appendices~\ref{sec:supp-experimental-details} and
\ref{sec:supp-ablation-details} specify the evaluation and ablation protocols.
Appendix~\ref{sec:supp-qualitative-extension} extends the long-horizon
qualitative comparison to all three backbones, and
Appendix~\ref{sec:supp-admission-montage} visualizes token admission events.

\section{Post-RoPE Attention Geometry}
\label{sec:supp-distance}

Let $\mathbf q_0,\mathbf k_0\in\mathbb{R}^{d_k}$ denote per-head query and key
vectors before positional encoding, and let $R_q$ and $R_k$ denote the RoPE
transforms actually applied to a current query and a nonlocal memory key at read
time. DensityKV operates on the resulting attention vectors
\begin{equation}
 \mathbf q=R_q\mathbf q_0,\qquad
 \mathbf k=R_k\mathbf k_0.
 \label{eq:supp-post-rope-vectors}
\end{equation}
Their pre-softmax scaled dot-product attention logit is
\begin{equation}
    \lambda(\mathbf q,\mathbf k)
    =\frac{\mathbf q^\top\mathbf k}{\sqrt {d_k}}.
    \label{eq:supp-logit}
\end{equation}
During memory attention, the evaluated backbones set the temporal RoPE coordinate
of nonlocal memory keys to zero while preserving their spatial coordinates.
DensityKV stores these same post-RoPE keys, so its retention geometry matches the
representation consumed by native attention. Future queries remain unknown when
the bank is updated.

\subsection{Attention-Logit Discrepancy}

The Cauchy--Schwarz inequality states that, for any vectors
$\mathbf a,\mathbf b\in\mathbb{R}^{d_k}$,
\begin{equation}
    |\mathbf a^\top\mathbf b|
    \leq\lVert\mathbf a\rVert_2\lVert\mathbf b\rVert_2,
    \label{eq:supp-cauchy-schwarz}
\end{equation}
with equality when $\mathbf a$ and $\mathbf b$ are linearly dependent. Let
$\Delta\mathbf k_{ij}=\mathbf k_i-\mathbf k_j$. Applying
Equation~\eqref{eq:supp-cauchy-schwarz} with $\mathbf a=\mathbf q$ and
$\mathbf b=\Delta\mathbf k_{ij}$ gives, for any query $\mathbf q$,
\begin{align}
 \left|\lambda(\mathbf q,\mathbf k_i)
       -\lambda(\mathbf q,\mathbf k_j)\right|
 &=\left|\frac{\mathbf q^\top\mathbf k_i
                   -\mathbf q^\top\mathbf k_j}{\sqrt {d_k}}\right| \notag\\
 &=\frac{1}{\sqrt {d_k}}
   \left|\mathbf q^\top(\mathbf k_i-\mathbf k_j)\right| \notag\\
 &\leq\frac{\lVert\mathbf q\rVert_2}{\sqrt {d_k}}
          \lVert\mathbf k_i-\mathbf k_j\rVert_2 .
 \label{eq:supp-bound}
\end{align}
For any nonzero $\mathbf q$, dividing by $\lVert\mathbf q\rVert_2$ and taking the
supremum gives
\begin{equation}
 \sup_{\mathbf q\neq \mathbf 0}
 \frac{\left|\lambda(\mathbf q,\mathbf k_i)
             -\lambda(\mathbf q,\mathbf k_j)\right|}
      {\lVert\mathbf q\rVert_2}
 =\frac{1}{\sqrt {d_k}}\lVert\mathbf k_i-\mathbf k_j\rVert_2 .
 \label{eq:supp-worst-case}
\end{equation}
The upper bound follows from Equation~\eqref{eq:supp-bound}; if
$\mathbf k_i\neq\mathbf k_j$, equality is attained by choosing any nonzero
$\mathbf q$ parallel to $\mathbf k_i-\mathbf k_j$. Thus Euclidean key distance
is exactly proportional to the worst-case normalized discrepancy of the
actual post-RoPE attention logits. Since RoPE transformations are orthogonal and
therefore invertible, expressing the query after RoPE does not weaken the
distribution-free supremum. Unlike a pre-RoPE surrogate, this geometry includes
the positional modulation used by the backbone for nonlocal memory access.

\subsection{Expected Key-Response Geometry and Cosine Distance}

Let $\mathcal Q$ denote the future post-RoPE query distribution. If this
distribution were known, a natural functional
distance would instead measure the expected squared logit discrepancy:
\begin{align}
 D_{\mathcal Q}(\mathbf k_i,\mathbf k_j)
 &=
 \mathbb{E}_{\mathbf q\sim\mathcal Q}\!\left[
   \left(\lambda(\mathbf q,\mathbf k_i)
        -\lambda(\mathbf q,\mathbf k_j)\right)^2
 \right] \notag\\
 &=
 \mathbb{E}_{\mathbf q\sim\mathcal Q}\!\left[
   \left(\mathbf q^\top\Delta\mathbf k_{ij}/\sqrt {d_k}\right)^2
 \right] \notag\\
 &=
 \frac{1}{d_k}
 \Delta\mathbf k_{ij}^{\top}
 \underbrace{
 \mathbb{E}_{\mathbf q\sim\mathcal Q}
 [\mathbf q\mathbf q^\top]}_{\Sigma_{\mathcal Q}}
 \Delta\mathbf k_{ij}.
 \label{eq:supp-query-distance}
\end{align}
This is a squared seminorm induced by the query second-moment matrix
$\Sigma_{\mathcal Q}$.

However, when future query statistics are unavailable or nonstationary, a
distribution-specific quadratic-form metric would require estimating and
continually tracking $\Sigma_{\mathcal Q}$. Under the isotropic query model
$\Sigma_{\mathcal Q}=\alpha I$, where $\alpha>0$ and $I$ is the identity
matrix, Equation~\eqref{eq:supp-query-distance} reduces to
\begin{align}
 D_{\mathcal Q}(\mathbf k_i,\mathbf k_j)
 &=\frac{1}{d_k}\Delta\mathbf k_{ij}^{\top}
       (\alpha I)\Delta\mathbf k_{ij} \notag\\
 &=\frac{\alpha}{d_k}
   \Delta\mathbf k_{ij}^{\top}\Delta\mathbf k_{ij} \notag\\
 &=\frac{\alpha}{d_k}\lVert\Delta\mathbf k_{ij}\rVert_2^2 \notag\\
 &=\frac{\alpha}{d_k}\lVert\mathbf k_i-\mathbf k_j\rVert_2^2.
 \label{eq:supp-isotropic}
\end{align}
Squared Euclidean distance is therefore the isotropic form of the
query-response geometry. Cosine distance instead compares only key
directions:
\begin{equation}
 d_{\mathrm{cos}}(\mathbf k_i,\mathbf k_j)
 =1-\frac{\mathbf k_i^\top\mathbf k_j}
          {\lVert\mathbf k_i\rVert_2\lVert\mathbf k_j\rVert_2}.
 \label{eq:supp-cosine}
\end{equation}
It is invariant to positive rescaling. In particular, for any
nonzero $\mathbf k$ and any $\gamma>0$,
\begin{equation}
 d_{\mathrm{cos}}(\mathbf k,\gamma\mathbf k)=0.
\end{equation}
Scaled dot-product attention is not invariant to the same operation:
\begin{equation}
 \left|\lambda(\mathbf q,\gamma\mathbf k)
       -\lambda(\mathbf q,\mathbf k)\right|
 =\frac{|\gamma-1|}{\sqrt {d_k}}|\mathbf q^\top\mathbf k|.
 \label{eq:supp-cosine-counterexample}
\end{equation}
For example, taking $\mathbf q=\mathbf k$ gives a logit discrepancy of
$|\gamma-1|\lVert\mathbf k\rVert_2^2/\sqrt {d_k}$ even though the cosine distance is
zero. Hence cosine distance cannot uniformly bound attention-logit
differences: keys that cosine treats as identical can produce
arbitrarily different logits as their relative norm changes.

This distinction matters because the evaluated autoregressive video
backbones do not constrain every per-head key to unit norm at the
attention input. Normalizing a larger hidden representation before
splitting it into heads does not fix the norm of each resulting
per-head key. Key norm therefore remains part of the model's native
attention response and should not be discarded by the bank-maintenance
geometry.

If a backbone did enforce unit-norm keys independently in every head,
then $\lVert\mathbf k_i\rVert_2=\lVert\mathbf k_j\rVert_2=1$. Expanding the squared
Euclidean distance gives
\begin{align}
 \lVert\mathbf k_i-\mathbf k_j\rVert_2^2
 &=(\mathbf k_i-\mathbf k_j)^\top
   (\mathbf k_i-\mathbf k_j) \notag\\
 &=\mathbf k_i^\top\mathbf k_i+\mathbf k_j^\top\mathbf k_j
   -2\mathbf k_i^\top\mathbf k_j \notag\\
 &=\lVert\mathbf k_i\rVert_2^2+\lVert\mathbf k_j\rVert_2^2
      -2\mathbf k_i^\top\mathbf k_j \notag\\
 &=2-2\mathbf k_i^\top\mathbf k_j \notag\\
 &=2\left(1-\frac{\mathbf k_i^\top\mathbf k_j}
 {\lVert\mathbf k_i\rVert_2\lVert\mathbf k_j\rVert_2}\right) \notag\\
 &=2d_{\mathrm{cos}}(\mathbf k_i,\mathbf k_j).
 \label{eq:supp-unitnorm-equivalence}
\end{align}
The first three equalities expand the quadratic form, the fourth applies the
unit-norm assumption, and the final two use the definition of cosine distance.
Thus, under independent per-head unit normalization, cosine and
squared Euclidean distance induce the same neighborhood ordering up to a
constant factor. Without such normalization, cosine distance discards norm
information that affects the actual post-RoPE dot-product response. DensityKV
consequently uses squared Euclidean key distance. Euclidean distance is exactly
proportional to the
worst-case normalized logit discrepancy in
Equation~\eqref{eq:supp-worst-case}; squaring it preserves the same neighborhood
ordering while enabling efficient pairwise evaluation. Keys determine memory
addressing, so values are excluded from the retention descriptor. The bank only
decides which key-addressable states remain available; this design does not
imply attention-output equivalence, and every retained value stays exactly
paired with its original key as the returned payload.

\section{Soft-Riesz Density and Parameterization}
\label{sec:supp-riesz}

\subsection{From Riesz Energy to a Smooth Crowding Score}

For post-RoPE keys separated by
$r=\lVert\mathbf k_i-\mathbf k_j\rVert_2$, a
classical Riesz interaction has the inverse-power form
\begin{equation}
    w_s^{\mathrm{Riesz}}(\mathbf k_i,\mathbf k_j)=r^{-s},
    \qquad s>0.
    \label{eq:supp-classical-riesz}
\end{equation}
Summing this interaction over all unordered pairs gives the classical
Riesz energy used to penalize locally crowded point configurations.
This form has the desired monotonicity---nearby
keys interact more strongly---but is singular when two keys coincide.
It also lacks an explicit scale for the distances encountered in a
particular attention head.

DensityKV therefore uses the regularized interaction
\begin{equation}
    w(\mathbf k_i,\mathbf k_j)
    =
    \left(
        \epsilon+
        \frac{r^2}{\sigma^2}
    \right)^{-p},
    \label{eq:supp-soft-riesz}
\end{equation}
where $\epsilon>0$ regularizes the interaction for near-duplicate keys,
$\sigma>0$ is the
neighborhood bandwidth, and $p>0$ controls the decay rate. The additive
$\epsilon$ removes the singularity, while $\sigma$ expresses distance in units
of the intended neighborhood width. To
make its relation to Equation~\eqref{eq:supp-classical-riesz}
explicit, rewrite it as
\begin{align}
    w(\mathbf k_i,\mathbf k_j)
    &=
    \left(\frac{r^2}{\sigma^2}\right)^{-p}
    \left(1+\frac{\epsilon\sigma^2}{r^2}\right)^{-p}
    \notag\\
    &=
    \sigma^{2p}r^{-2p}
    \left(1+\frac{\epsilon\sigma^2}{r^2}\right)^{-p}.
    \label{eq:supp-riesz-asymptotic}
\end{align}
For $r^2\gg\epsilon\sigma^2$, the final factor approaches one, so the
interaction has the same far-field decay as a Riesz kernel with order
$s=2p$. For two distinct indexed entries whose key vectors coincide, however,
\begin{equation}
    w(\mathbf k_i,\mathbf k_j)=\epsilon^{-p}
    \qquad(i\neq j,\ \mathbf k_i=\mathbf k_j),
\end{equation}
which is finite. We use the term \emph{Soft-Riesz} to denote this
regularized inverse-power interaction.

For a finite indexed key multiset
$\mathcal A=(\mathbf k_j)_{j=1}^{|\mathcal A|}$, the potential of entry $i$ is
\begin{equation}
    \Phi_{\mathcal A}(i)=
    \sum_{\substack{j=1\\j\neq i}}^{|\mathcal A|}
    w(\mathbf k_i,\mathbf k_j),
\end{equation}
and the pairwise energy is
\begin{equation}
    E(\mathcal A)
    =
    \frac{1}{2}\sum_{i=1}^{|\mathcal A|}
    \Phi_{\mathcal A}(i)
    =
    \frac{1}{2}\sum_{i=1}^{|\mathcal A|}
    \sum_{\substack{j=1\\j\neq i}}^{|\mathcal A|}
    w(\mathbf k_i,\mathbf k_j).
    \label{eq:supp-pairwise-energy}
\end{equation}
The factor $1/2$ removes the double counting in the potential sum.
Appending a new indexed entry $\mathbf k_x$ changes the energy by
\begin{equation}
    E(\mathcal A\uplus\{\mathbf k_x\})-E(\mathcal A)
    =
    \sum_{j=1}^{|\mathcal A|}w(\mathbf k_x,\mathbf k_j)
    =
    \Phi_{\mathcal A}(\mathbf k_x),
    \label{eq:supp-insertion-energy}
\end{equation}
where $\uplus$ denotes indexed multiset insertion and
$\Phi_{\mathcal A}(\mathbf k_x):=\sum_{j=1}^{|\mathcal A|}
w(\mathbf k_x,\mathbf k_j)$ denotes the candidate's potential against the
existing entries.
Thus the potential is exactly the marginal interaction energy of an
insertion: a large value means that the candidate interacts strongly with an
already crowded neighborhood. This ``density'' is an
unnormalized operational crowding score, not a probability density
estimate. DensityKV only compares scores within the same layer and
head, where the geometry and bank capacity are shared. The pairwise
energy is used only to motivate the local potential; the online update
does not compare global set energies or perform group-energy replacement.

\subsection{Parameterization}

The regularizer and bandwidth can be separated as
\begin{equation}
    \left(
        \epsilon+\frac{r^2}{\sigma^2}
    \right)^{-p}
    =
    \epsilon^{-p}
    \left(
        1+\frac{r^2}{\epsilon\sigma^2}
    \right)^{-p}.
    \label{eq:supp-riesz-scale}
\end{equation}
Consequently, $\epsilon^{-p}$ is a common positive multiplier and
$\sigma\sqrt{\epsilon}$ is the effective bandwidth. A common multiplier does
not change density rankings or normalized growth ratios when the numerical
floor is inactive; strict scale equivalence would additionally require scaling
$\delta$ by the same factor. We fix $\epsilon=1$, which sets the maximum
pairwise contribution to one and leaves $\sigma$ as the sole distance-scale
parameter in our experiments.

We set $p=2$, giving a smooth inverse-quartic far-field decay. This choice
emphasizes nearby interactions without imposing a hard nearest-neighbor
boundary. The same exponent is used in every layer, head, backbone, and
generation length.

Finally, we calibrate the bandwidth to the key dimension using
\begin{equation}
    \sigma=\sqrt{d_k/2}.
    \label{eq:supp-bandwidth}
\end{equation}
For intuition about this scale, if two keys have norms
near $\sqrt {d_k}$ and cosine similarity $\cos\theta$, then
\begin{equation}
    \frac{\lVert\mathbf k_i-\mathbf k_j\rVert_2^2}{\sigma^2}
    \approx
    \frac{2d_k(1-\cos\theta)}{d_k/2}
    =
    4(1-\cos\theta).
    \label{eq:supp-bandwidth-intuition}
\end{equation}
This relation is only a bandwidth calibration; DensityKV still uses native,
non-unit-normalized post-RoPE keys and therefore retains their norm information.
All evaluated backbones have $d_k=128$, for which
Equation~\eqref{eq:supp-bandwidth} gives $\sigma=8$. We use
$(\epsilon,p,\sigma)=(1,2,8)$ by default. The main paper varies $\sigma$ and
$p$; $\epsilon$ remains fixed because varying it jointly rescales the kernel
amplitude and effective bandwidth.

\section{Complete Density-Guided KV Bank Update}
\label{sec:supp-update-algorithm}

Algorithm~\ref{alg:supp-densitykv} gives the complete online update for one
Transformer layer with $H$ attention heads and per-head budget $M$. For head
$h$, $\mathcal C^h$ contains all $N$ candidate K/V pairs from one finalized
generation block, and
$\bar{\mathbf k}_i^h$ denotes the post-RoPE key of retained state $i$.
In our evaluated configuration, one block contributes $N=4{,}680$ candidates.
Within each head, the full set participates in a single ordering, feasibility
test, and bank commit; it is not divided into multiple policy updates. Each
head maintains its own candidate ordering, eviction set, and exact K/V pairs,
while all heads in the layer share only the admitted prefix length. Let
$\mathcal K(\mathcal B^h)$ denote the indexed key multiset in bank
$\mathcal B^h$. For retained state $i$, $\rho_i^h=
\Phi_{\mathcal K(\mathcal B^h)}(i)$ is its cached current density,
$b_i^h$ is its frozen insertion-density baseline, $\tau>1$ is the
allowed density-growth factor, and $\delta>0$ is the baseline floor.

\paragraph{Ordered candidates and shared feasibility.}
For a nonempty bank with occupancy $m$, candidate $x$ in head $h$ is scored by
its mean-normalized disturbance over the retained bank,
\begin{equation}
s_h(x)=\frac{1}{m}\sum_{i\in\mathcal B^h}
\frac{w(\bar{\mathbf k}_i^h,\mathbf k_x^h)}{b_i^h}.
\label{eq:supp-candidate-score}
\end{equation}
This measures the average insertion-normalized disturbance that the candidate
would impose across retained neighborhoods.
Stable sorting by increasing score gives a head-specific permutation $\pi_h$
and prefixes
$\mathcal P_r^h=((\mathbf k_{\pi_h(j)}^h,\mathbf
v_{\pi_h(j)}^h))_{j=1}^{r}$, with $\mathcal P_0^h=\varnothing$. For any tested
prefix length $r$, define the projected density and its insertion-normalized
ratio for historical state $i$ as
\begin{align}
 \widetilde{\rho}_i^h(r)
 &=
 \rho_i^h+
 \sum_{j=1}^{r}
 w(\bar{\mathbf k}_i^h,\mathbf k_{\pi_h(j)}^h),
 \label{eq:supp-projected-density}\\
 g_i^h(r)
 &=
 \frac{\widetilde{\rho}_i^h(r)}{b_i^h}.
 \label{eq:supp-density-growth}
\end{align}
Because synchronized admission keeps the occupancy
$m=|\mathcal B^h|$ equal across heads, admitting $r$ candidates requires
$e_r=\max(0,m+r-M)$ evictions. Let
$c_r^h=|\{i\in\mathcal B^h:g_i^h(r)\geq\tau\}|$ count the historical states
that would violate the growth bound. A prefix is feasible for head $h$ when
$c_r^h\leq e_r$, and the layer selects the largest length feasible for every
head. In the main paper's notation, $g_i^h(r)$ and $c_r^h$ are the
expanded forms of $r_i^h(\mathcal B^h,\mathcal P_r^h)$ and $v_r^h$ in
Equations (8) and (9), respectively.
Thus all heads admit the same number of states, although the admitted token
identities may differ across heads. The maintained growth invariant makes
$r=0$ feasible; if no positive prefix is feasible, the bank may remain below
capacity.

\begin{algorithm*}[!t]
\small
\caption{DensityKV update for one Transformer layer}
\label{alg:supp-densitykv}
\begin{algorithmic}[1]
\REQUIRE Per-head banks $\{\mathcal B^h\}_{h=1}^{H}$ with cached densities
$\rho_i^h$, frozen baselines $b_i^h$, common occupancy
$m=|\mathcal B^1|=\cdots=|\mathcal B^H|\leq M$, and budget $M$;
one full-block candidate set $\{\mathcal C^h\}_{h=1}^{H}$; threshold $\tau$;
baseline floor $\delta$
\IF{$m=0$}
  \STATE Set $r^\star\leftarrow\min(N,M)$
  \FOR{each head $h$}
    \STATE Initialize $\mathcal B^h$ with the first $r^\star$
           source-ordered candidates
    \STATE For every admitted state $i$, set
           $\rho_i^h\leftarrow\Phi_{\mathcal K(\mathcal B^h)}(i)$ and
           $b_i^h\leftarrow\max(\rho_i^h,\delta)$
  \ENDFOR
\ELSE
  \FOR{each head $h$}
    \STATE Stably sort candidates by increasing mean-normalized score
           $s_h(x)$ from Equation~\eqref{eq:supp-candidate-score},
           breaking ties by source order, to form prefixes
           $\{\mathcal P_r^h\}_{r=0}^{\min(N,M)}$
    \FOR{each tested prefix length $0\leq r\leq\min(N,M)$}
      \STATE Set the required eviction count
             $e_r\leftarrow\max(0,m+r-M)$
      \STATE Project every retained state's density ratio
             $g_i^h(r)$ by Equation~\eqref{eq:supp-density-growth}
      \STATE Count violators
             $c_r^h\leftarrow|\{i:g_i^h(r)\geq\tau\}|$
    \ENDFOR
  \ENDFOR
  \STATE Select the largest shared prefix length
         $r^\star$ satisfying $c_r^h\leq e_r$ for every head
  \FOR{each head $h$}
    \STATE Initialize $\mathcal R^h$ with all historical states satisfying
           $g_i^h(r^\star)\geq\tau$
    \STATE Add states from $\mathcal B^h\setminus\mathcal R^h$ in decreasing
           $\widetilde{\rho}_i^h(r^\star)$, breaking ties arbitrarily, until
           $|\mathcal R^h|=e_{r^\star}$
    \STATE Commit
           $\mathcal B^h\leftarrow
           (\mathcal B^h\setminus\mathcal R^h)
           \uplus\mathcal P_{r^\star}^h$
    \STATE Update the cached density $\rho_i^h$ of every surviving historical
           state exactly by adding interactions with admitted states and
           subtracting interactions with evicted states
    \STATE For each newly admitted state $i$, set
           $\rho_i^h\leftarrow\Phi_{\mathcal K(\mathcal B^h)}(i)$ and
           $b_i^h\leftarrow\max(\rho_i^h,\delta)$;
           keep every surviving historical baseline unchanged
  \ENDFOR
\ENDIF
\RETURN Per-head banks of exact K/V pairs $\{\mathcal B^h\}_{h=1}^{H}$
\end{algorithmic}
\end{algorithm*}

\paragraph{Insertion hysteresis.}
All candidates admitted from one finalized generation block are committed
simultaneously. Their mutual interactions do not participate in the
shared-feasibility test; instead, they are
included when the new states' insertion baselines are computed after the
commit. Candidates from the same block therefore do not reject one another
during that update. The resulting baselines define their neighborhoods at
admission, and only interactions introduced by later generation blocks count
as post-admission growth that can trigger eviction. DensityKV therefore
constrains \emph{post-admission density growth}, not absolute bank density. The
within-block competition ablation in the main paper instead uses sequential
admission, in which earlier candidates immediately affect later admission
decisions.

\paragraph{Exact density maintenance.}
For a surviving historical state $i$, the commit updates its cached density as
\begin{equation}
 \rho_i^{h,\mathrm{new}}
 =
 \rho_i^h+
 \sum_{\mathbf c\in\mathcal K(\mathcal P_{r^\star}^h)}
 w(\bar{\mathbf k}_i^h,\mathbf c)
 -
 \sum_{\mathbf k_j\in\mathcal K(\mathcal R^h)}
 w(\bar{\mathbf k}_i^h,\mathbf k_j).
 \label{eq:supp-density-update}
\end{equation}
For each newly admitted state, its density is computed against the final bank,
including the other co-admitted candidates, and that value is frozen, subject
to the floor $\delta$, as its insertion baseline. Consequently, no interaction
matrix among previously retained states is rebuilt after each update.

\paragraph{Decision granularity versus computational tiling.}
The policy above executes exactly one bank update for each complete candidate
block. Implementations may tile the candidate--bank interaction matrix to bound
temporary workspace. These tiles only accumulate the exact scores, prefix
contributions, and density ratios required by the same full-block decision.
No bank state or insertion baseline is updated between tiles, and tile
boundaries have no algorithmic role.
An implementation may cache the candidate--bank interaction matrix and permute
its columns after sorting, or recompute the same interactions in a second tiled
pass. Column-wise cumulative sums then produce all prefix contributions without
recomputing them separately for each tested prefix.
Computing mutual interactions among the $r^\star$ co-admitted states costs
$O((r^\star)^2d_k)$, while subtracting interactions with
$e_{r^\star}$ evicted states costs $O(Me_{r^\star}d_k)$. Since
$r^\star,e_{r^\star}\leq N\leq M$, both terms are covered by the
$O(NMd_k)$ interaction cost reported in the main paper.

\paragraph{Invariant after commit.}
The shared-feasibility condition guarantees that the number of historical
states whose projected ratios reach $\tau$ does not exceed the eviction budget.
Any surviving historical state had a projected ratio below $\tau$ before
eviction; because $w\geq0$, removing other states can only decrease its density.
Every newly
admitted state stores
$b_i=\max(\rho_i,\delta)$ after the final bank is formed, so its initial ratio
is at most one and therefore below $\tau$. The committed bank consequently
restores the density-growth invariant for all retained states.
\FloatBarrier

\section{Experimental Details}
\label{sec:supp-experimental-details}

\paragraph{Backbones and prompts.}
All DensityKV experiments use frozen autoregressive video generators built from
Wan2.1-T2V-1.3B: LongLive, Self-Forcing, and Causal-Forcing. DensityKV is
applied only at inference time and does not fine-tune the backbone. The main
comparison follows the LongLive-RAG protocol with all 128 MovieGenBench prompts
refined by Qwen2.5-7B-Instruct. We report 30-, 60-, and 120-second generations.
Our DensityKV runs use seed 0 for every prompt.
For each prompt and backbone, DensityKV is run once to 120 seconds; the 30- and
60-second videos are packet-preserving truncations and therefore exact encoded
prefixes of the same trajectory.
Controlled ablations use a fixed 16-prompt subset from the same prompt pool and
the 60-second endpoint. The subset uses zero-based prompt indices 0, 8, 17, 25,
34, 42, 51, 65, 68, 76, 85, 93, 102, 110, 119, and 127 in the released refined
prompt list, with seed 0 for every run. Parameter sweeps retain the default
DensityKV policy, and named policy rows vary one update component while holding
the others fixed. Within-block competition changes admission granularity and is
not a strict one-factor ablation. The auxiliary geometry suite keeps
mean-normalized ordering while varying the retention descriptor.

\paragraph{Memory budget and temporal alignment.}
The evaluated Wan latent grid has 1,560 spatial tokens per frame for each
attention head. DensityKV and LongLive-RAG use the same upper bound on
attention-visible nonlocal historical capacity, corresponding to six
latent-frame equivalents per head, or $6\times1{,}560=9{,}360$ KV states.
Their rollouts also keep the same explicit
first-frame sink and local window; other baselines follow the configurations
reported by LongLive-RAG. Thus, DensityKV's default visible temporal
context comprises a one-frame sink ($1{,}560$ tokens), a five-frame local
window ($7{,}800$ tokens), and up to six nonlocal frame-equivalents
($9{,}360$ tokens). DensityKV's bank may remain below this maximum early in a
video, so the matched quantity is capacity rather than exact instantaneous
occupancy.

\paragraph{Persistent-storage accounting.}
Let $L$, $H$, $d_k$, and $d_v$ denote the numbers of layers and heads and the
per-head key and value dimensions. For $n$ retained states per head, the BF16
K/V payload occupies
\begin{equation}
 S_{\mathrm{KV}}(n)
 =LHn(d_k+d_v)\times 2\ \text{bytes}.
\end{equation}
With $L=30$, $H=12$, $d_k=d_v=128$, and $n=M=9{,}360$, the DensityKV
bank payload is 1.61 GiB. Two FP32 density scalars and int64/int32 provenance
fields per state raise the bank total to 1.67 GiB. The explicit sink and five-frame
local window contain another $6\times1{,}560=9{,}360$ K/V states per head,
yielding 3.28 GiB of total persistent temporal-state storage. LongLive-RAG includes all
historical BF16 K/V states and retrieval descriptors; Deep Forcing includes its
12-frame BF16 K/V cache, four-frame BF16 query buffer with per-head dimension
$d_k=128$, and metadata. Model parameters and
layer-local temporary workspace are excluded for every method.
For the plotted horizons, LongLive-RAG stores 120, 240, and 480 latent frames,
respectively, with one 1,024-dimensional BF16 retrieval descriptor per frame.
The Deep Forcing total includes 0.01 GiB of persistent count, index, and mask
metadata in addition to its K/V cache and query buffer.

\paragraph{Default DensityKV configuration.}
Unless varied in an ablation, DensityKV uses post-RoPE keys for density
computation,
keeps values exactly paired with their original keys, and maintains one
independent bank for each layer--head pair. The default parameters are
$M=9{,}360$, $\tau=2.0$, $\sigma=8$, $p=2$, $\epsilon=1$, and
$\delta=10^{-6}$. One finalized generation block contributes
$N=4{,}680$ candidates per head, corresponding to three latent frames. The
whole block participates in a single sorting operation, feasibility test, and
commit decision; any
matrix tiling is only an implementation detail and does not create intermediate
policy updates. The bank is updated only during a fully denoised $t=0$ cache pass,
when a finalized block leaves the five-frame local window; denoising forward
passes at $t>0$ only read the retained K/V bank. The default uses mean-normalized
candidate ordering, frozen insertion baselines, mandatory-then-densest eviction,
and a prefix length shared across the heads of each layer.

\paragraph{Metrics.}
We follow the LongLive-RAG VBench-Long reporting protocol and use the same six
VBench dimensions: Subject Consistency, Background Consistency, Motion
Smoothness, Dynamic Degree, Aesthetic Quality, and Imaging Quality. Higher is
better for all six metrics. Main-table ranks use the displayed two-decimal
values with midranks for ties. Each group Avg. Rank averages its six metric
ranks, and the overall value spans all 54 backbone--horizon--metric cells.
Ablation scores are unweighted three-backbone macro means. Their Avg. Rank
averages competition ranks over the six displayed means, with ties ranked as
$1,1,3$, within the three settings of each parameter sweep, the seven policy
variants, or the four geometry variants, respectively.

\section{Ablation Details}
\label{sec:supp-ablation-details}

\paragraph{Parameter sweeps.}
The parameter ablations vary one scalar while keeping the rest of the default
configuration fixed. The threshold sweep changes the allowed post-admission
density growth factor $\tau$. The capacity sweep changes the per-head bank
limit $M$. The bandwidth sweep changes the Soft-Riesz neighborhood scale
$\sigma$. The exponent sweep changes the inverse-power decay $p$. We keep
$\epsilon=1$ because changing $\epsilon$ mainly rescales the effective bandwidth
as discussed in Appendix~\ref{sec:supp-riesz}.

\paragraph{Candidate ordering.}
The default orders candidates by their mean-normalized disturbance: the
Soft-Riesz interaction with every retained key is divided by that key's frozen
baseline, and the resulting ratios are averaged over the bank. Lower scores are
considered earlier.
The \emph{Source-order candidates} ablation instead uses the original
spatiotemporal order.

\paragraph{Density baseline.}
Each admitted key stores its insertion-time density as a fixed local baseline.
The \emph{Refresh density baselines} ablation recomputes it after every update,
testing whether the admission neighborhood should remain the reference or track
the evolving bank.

\paragraph{Eviction rule.}
The default first evicts states whose projected density ratio crosses the
threshold, then completes the eviction set with unselected states in decreasing
unnormalized projected density $\widetilde{\rho}_i^h(r^\star)$.
\emph{Densest-only eviction} uses only this ordering; \emph{Mandatory + source
eviction} preserves threshold-crossing removals but fills additional slots in
source order.

\paragraph{Cross-head synchronization.}
Each head computes its own candidate order and eviction set. The default selects
the largest prefix length feasible for every head in a layer, so all heads admit
the same number of candidates per update while admitting different token
identities. The \emph{Independent per-head} ablation lets each attention head
choose its own prefix length.

\paragraph{Within-block competition.}
The default commits the selected prefix from a finalized block in one update and
includes mutual candidate interactions only in the new insertion baselines.
\emph{Within-block
competition} instead uses sequential density-gated admission, so earlier
candidates immediately affect later ones. It also changes admission granularity
and is therefore a mechanism comparison rather than a strict one-factor
ablation.

\paragraph{Retention geometry.}
The geometry suite compares pre- and post-RoPE $K$ and $[K;V]$ descriptors
under the same mean-normalized ordering as the final policy.
Post-RoPE keys are the vectors consumed by nonlocal attention; memory addressing
is determined by Q/K logits rather than values, which remain paired with their
keys. Adding $V$ improves pre-RoPE
Avg. Rank from 2.33 to 2.00 but worsens post-RoPE rank from 1.83 to 3.17.

\clearpage
\onecolumn

\begin{figure}[!p]
\section{Extended Long-Horizon Qualitative Comparisons}
\label{sec:supp-qualitative-extension}
\centering
\makebox[\textwidth][c]{\includegraphics[width=1.02\textwidth,height=0.92\textheight,keepaspectratio]{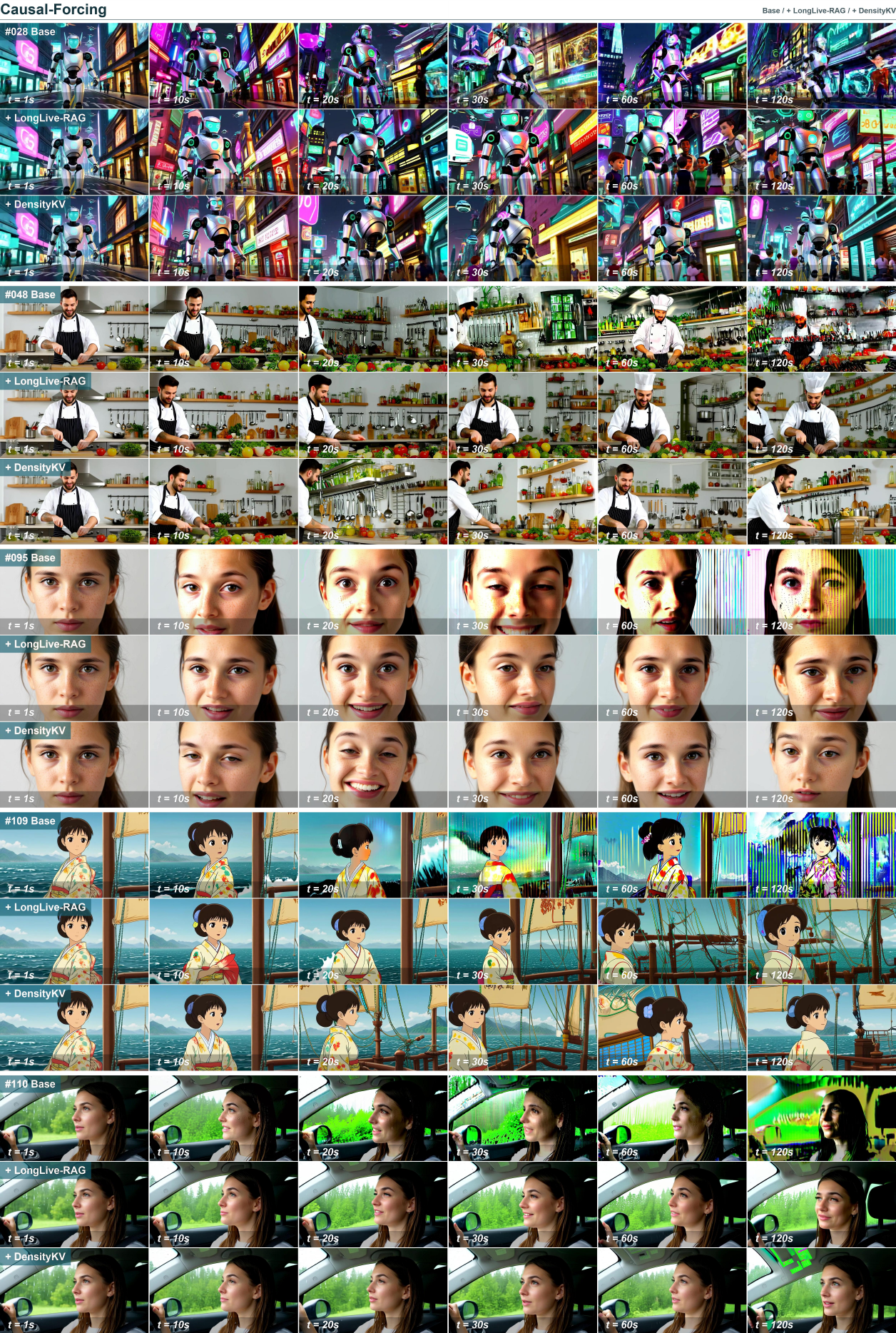}}
\caption{Extended qualitative comparison.
Causal-Forcing on MovieGenBench prompts \#028, \#048, \#095, \#109, and
\#110. Rows: Native, LongLive-RAG, and post-RoPE K-only DensityKV. Columns:
1, 10, 20, 30, 60, and 120 seconds from matched trajectories.}
\label{fig:supp-figure3-extension-a}
\end{figure}
\clearpage

\begin{figure}[!p]
\centering
\makebox[\textwidth][c]{\includegraphics[width=1.02\textwidth,height=0.97\textheight,keepaspectratio]{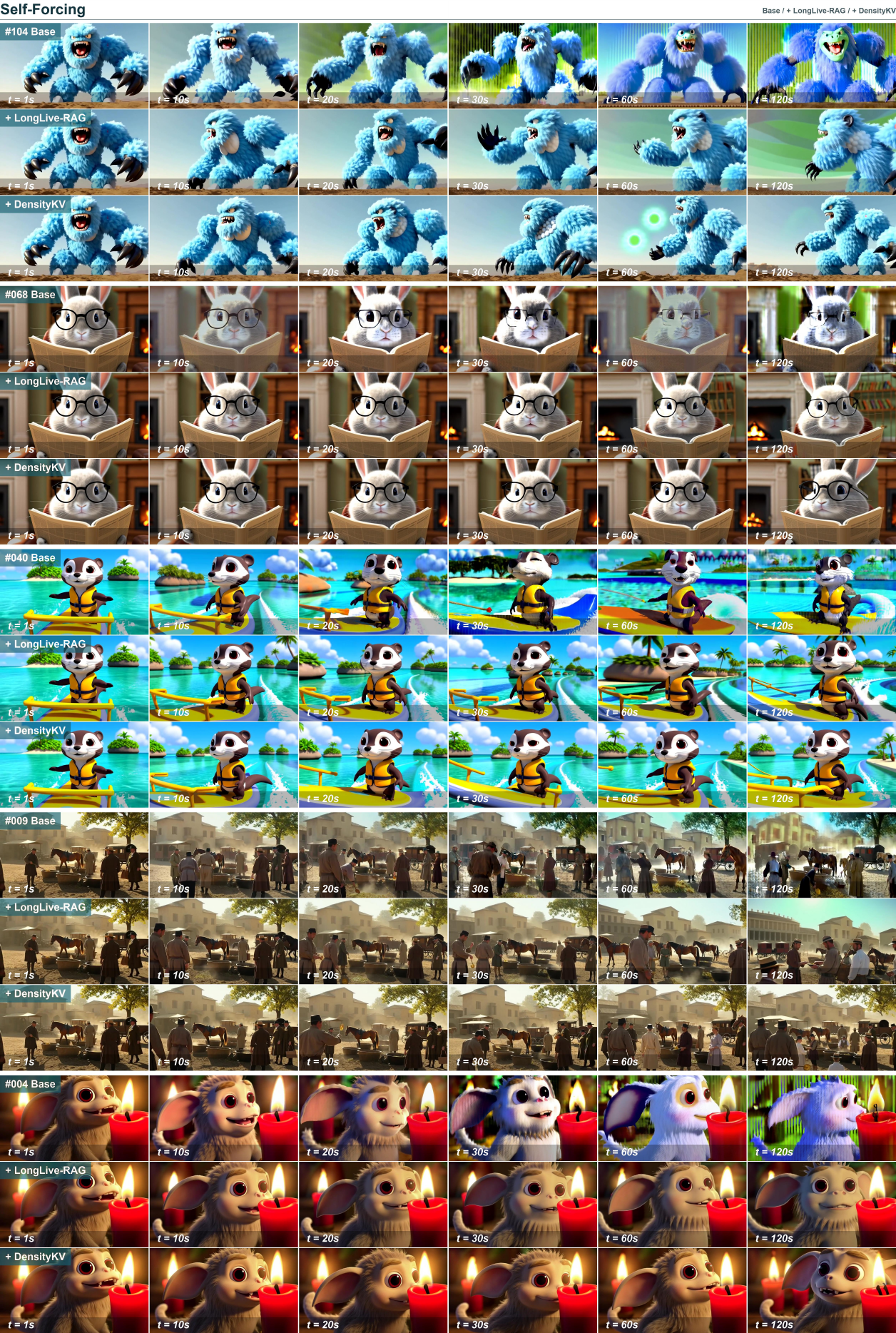}}
\caption{Self-Forcing on MovieGenBench prompts \#104, \#068, \#040, \#009, and
\#004. Rows: Native, LongLive-RAG, and DensityKV. Columns: 1, 10, 20, 30, 60,
and 120 seconds.}
\label{fig:supp-figure3-extension-b}
\end{figure}
\clearpage

\begin{figure}[!p]
\centering
\makebox[\textwidth][c]{\includegraphics[width=1.02\textwidth,height=0.97\textheight,keepaspectratio]{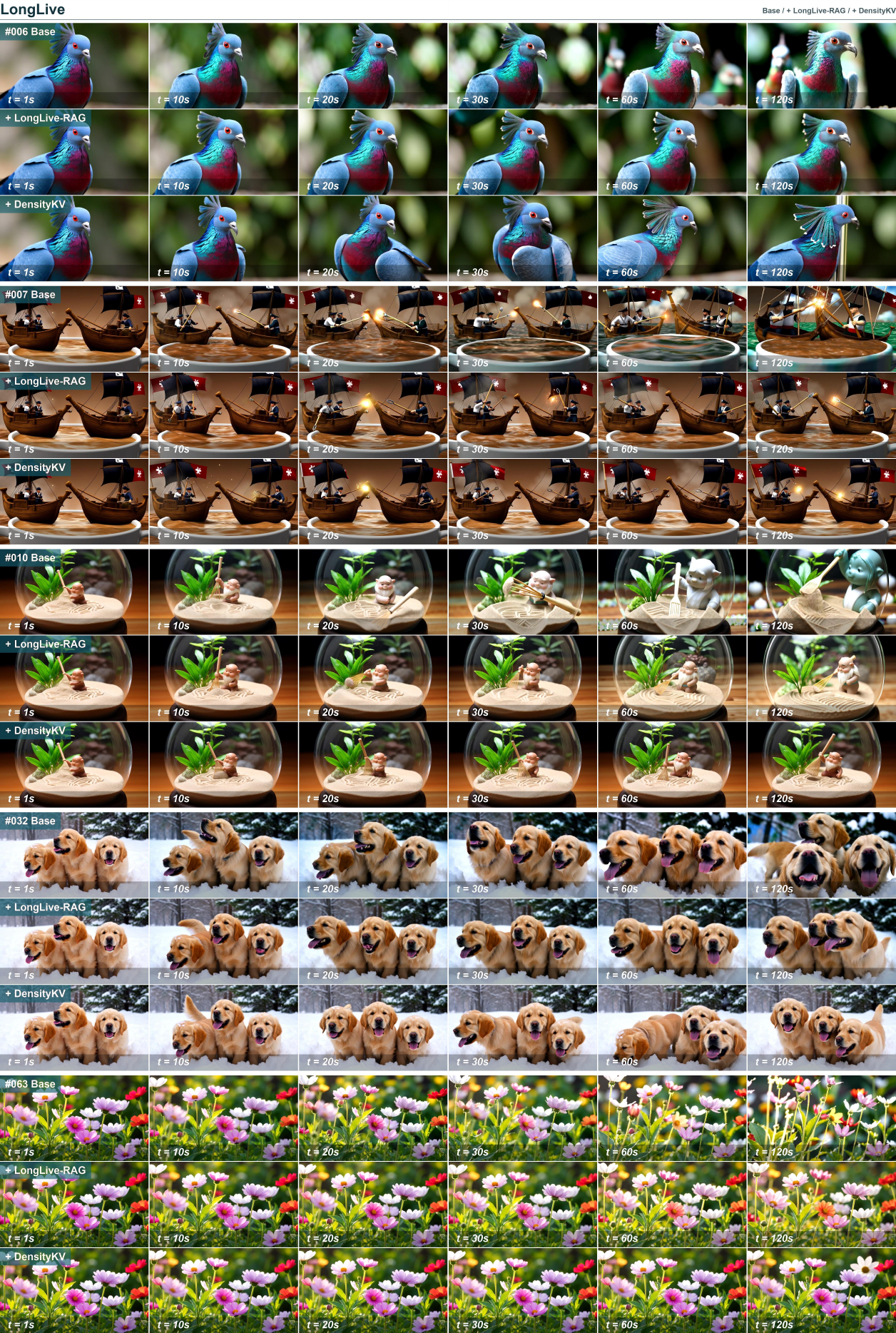}}
\caption{LongLive on MovieGenBench prompts \#006, \#007, \#010, \#032, and
\#063. Rows: Native, LongLive-RAG, and DensityKV. Columns: 1, 10, 20, 30, 60,
and 120 seconds.}
\label{fig:supp-figure3-extension-c}
\end{figure}
\clearpage

\section{Token Admission Frequency and Layerwise Coverage}
\label{sec:supp-admission-montage}

Figure~\ref{fig:supp-admission-montage} resolves the admission events across all
30 layers and 12 attention heads instead of collapsing them into a binary
union.

\begin{figure}[!ht]
\centering
\includegraphics[width=\textwidth]{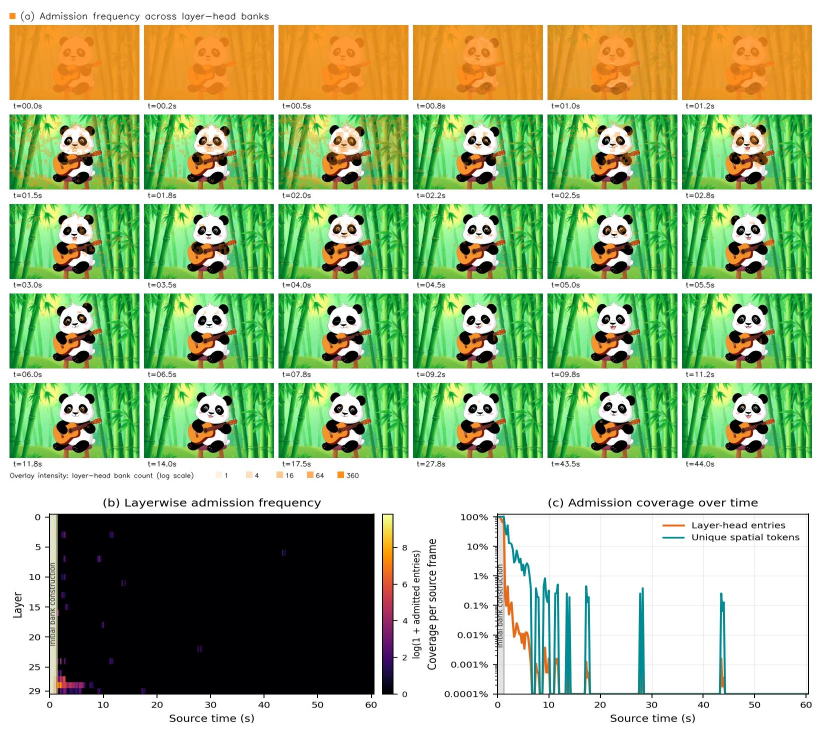}
\caption{Admission analysis for the Causal-Forcing panda-guitar diagnostic
(seed 13). (a) Orange intensity gives the number of layer--head banks that
admit each spatial token, on a logarithmic scale. (b) Admission counts
aggregated by layer and source time. (c) Fractions of possible layer--head
entries and unique spatial positions admitted from each source frame. The gray
region in (b)--(c) marks initial bank construction from source frames 1--6,
whose timestamps span 0--1.25 seconds at 4 fps.}
\label{fig:supp-admission-montage}
\end{figure}

\paragraph{Interpretation.}
During initial bank construction, the first six source frames (0--1.25 seconds
at 4 fps) contribute 99.77\% of the retained layer--head entries in this trace.
This concentration primarily reflects the matched capacity
$M=9{,}360=6\times1{,}560$, rather than evidence that DensityKV identifies the
beginning of the video as intrinsically more important. After initialization,
admissions become sparse and layer-specific, showing that the density-growth
test can still admit under-covered states from later content, including
isolated admissions between roughly 10 and 45 seconds. The trace contains
2,990,532 admitted layer--head entries and 12,602 unique admitted source-token
positions, with no subsequent eviction because occupancy remains below the
capacity bound; therefore a final-retention mask would exactly duplicate panel
(a).

\fi

\end{document}